\documentclass{article}

\usepackage{PRIMEarxiv}
\usepackage{tikz}
\usepackage{caption} 
\usepackage{algorithm}
\usepackage{algpseudocode}
\usepackage{booktabs}
\usepackage{amssymb}
\usepackage{amsthm}
\usepackage[utf8]{inputenc}
\usepackage{textgreek}
\usepackage{amsfonts}
\usepackage{multirow}
\usepackage[table]{xcolor}
\definecolor{oraclerow}{RGB}{225,235,245}
\newcommand{\cmark}{\ensuremath{\checkmark}}
\newcommand{\xmark}{\ensuremath{\times}}
\usepackage{amsmath}
\usepackage{comment}
\excludecomment{draftonly}   

\usepackage{booktabs}   
\usepackage{multirow}   
\usepackage[table]{xcolor} 
\usepackage{pifont}

\usepackage{amsmath}
\usepackage{amssymb} 
\usepackage{amsthm}
\usepackage{algorithm}
\usepackage{algpseudocode}
\usepackage{dsfont}
\usepackage{tikz} 
\usepackage{xcolor}
\usepackage{soul}

\usepackage[table]{xcolor} \definecolor{oraclerow}{RGB}{240,240,240}

\newtheorem{assumption}{Assumption}
\newtheorem{proposition}{Proposition}
\newtheorem{theorem}{Theorem}
\newtheorem{corollary}{Corollary}
\theoremstyle{remark}

\usepackage{newfloat}

\usepackage[utf8]{inputenc} 
\usepackage[T1]{fontenc}    
\usepackage{hyperref}       
\usepackage{url}            
\usepackage{booktabs}       
\usepackage{amsfonts}       
\usepackage{nicefrac}       
\usepackage{microtype}      
\usepackage{lipsum}
\usepackage{fancyhdr}       
\usepackage{graphicx}       
\graphicspath{{media/}}     

\title{ReCAP: Sequential Concept Acquisition with Guarantees under Imperfect Concept Predictors
}

\author{
  Md Nahiduzzaman \\
  School of Computing Technologies \\
  RMIT University \\
  Melbourne, Australia \\
  \texttt{nahiduzzaman@ece.ruet.ac.bd} 
  \And
  Steven Korevaar \\
  School of Computing Technologies \\
  RMIT University \\
  Melbourne, Australia \\
  \texttt{steven.korevaar@rmit.edu.au}
  \And
  Alireza Bab-Hadiashar \\
  School of Engineering \\
  RMIT University \\
  Melbourne, Australia \\
  \texttt{alireza.bab-hadiashar@rmit.edu.au}
  \And
  Ruwan Tennakoon \\
  School of Computing Technologies \\
  RMIT University \\
  Melbourne, Australia \\
  \texttt{ruwan.tennakoon@rmit.edu.au}
}

\begin{document}
\maketitle

\begin{abstract}
Sequential concept acquisition enables interpretable diagnosis from a compact set of human-understandable findings. Such explanations are useful only if clinicians can trust them, yet most existing methods treat predicted concepts as correct, so a correct diagnosis can rest on incorrectly predicted evidence. We propose Reliable Concept Acquisition with Pruning (ReCAP), which reasons over what the concept predictor reports, including errors and abstentions, and replaces conditional information gain with a label-discriminating objective that separates cases requiring different diagnoses. We prove this objective adaptively monotone and adaptively submodular, yielding a $(1-1/e)$ guarantee for greedy fixed-budget acquisition that degrades boundedly under learned estimates, and combine it with abstention-aware acquisition and diagnosis-preserving pruning to keep explanations short and reliable. Across five medical imaging datasets, ReCAP reduces the share of correct diagnoses supported by at least one incorrectly predicted concept from 27.1\% to 9.6\% relative to the strongest baseline, while using fewer than half as many concepts and achieving higher diagnostic accuracy. 
\end{abstract}


\section{Introduction}
\label{sec:intro}

In safety-critical medical settings, a correct diagnosis is not sufficient on its
own; clinicians and regulators also need to trust the reasoning behind it, which
is difficult when that reasoning comes from a black-box network. Concept
Bottleneck Models (CBMs)~\cite{koh2020concept} address this by introducing an
intermediate layer of human-understandable concepts, so that predictions can be
explained through clinically meaningful features. To better reflect clinical
reasoning, recent sequential methods acquire these concepts adaptively, one
query at a time~\cite{chattopadhyay2022interpretable, chattopadhyayvariational,
nahiduzzaman2026uncertainty}, basing the diagnosis on the smallest necessary
subset of evidence rather than evaluating every concept in the vocabulary.
\begin{figure}[!t]
\centering
\includegraphics[width=\columnwidth]{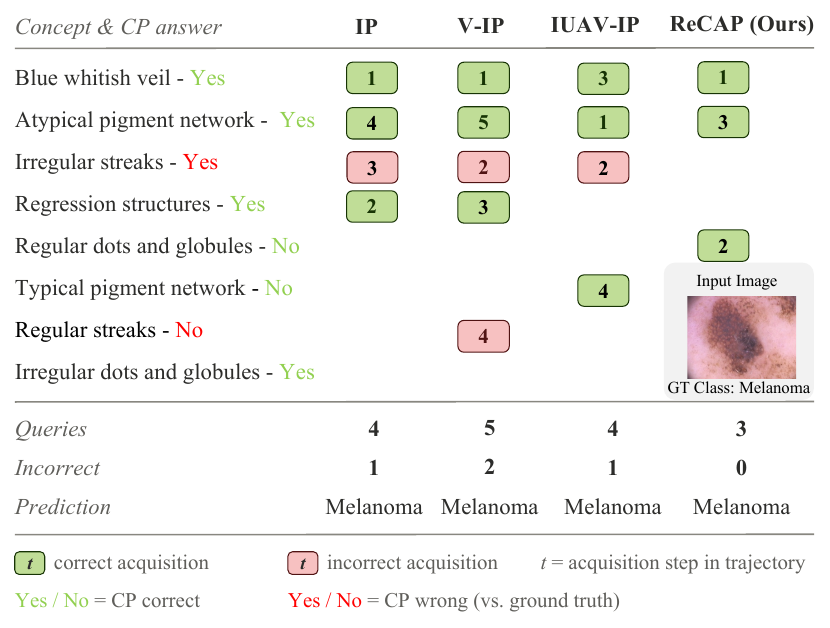}
\caption{
A correct diagnosis can rest on an untrustworthy explanation.
For this PH2 image, all four methods correctly predict Melanoma. However, IP~\cite{chattopadhyay2022interpretable}, V-IP~\cite{chattopadhyayvariational} and the uncertainty-aware IUAV-IP~\cite{nahiduzzaman2026uncertainty} acquire concepts that the concept predictor got wrong (red). ReCAP reaches the same diagnosis with fewer queries and no incorrectly predicted concepts.
}
\label{fig:motivation}
\end{figure}

Sequential acquisition raises two questions: how few concepts suffice, and whether the acquired values can be trusted. Together, these reflect the clinical need for an explanation concise enough to review, in which every asserted finding agrees with the image so that clinicians can verify the diagnostic rationale. Most sequential methods target the first question, selecting concepts as if the
concept predictor (CP) always returns the true concept.
In practice, CP errors are often \emph{instance-specific}. A concept that is reliable on average may be mispredicted for a particular image due to poor image quality, imaging artifacts, or atypical disease presentation. Consequently, a concept acquisition trajectory can reach the correct diagnosis while relying on incorrectly predicted concepts, producing a correct answer supported by false evidence (Fig.~\ref{fig:motivation}). This is not an isolated
case: for the strongest baseline, 27.07\% of correct diagnoses, averaged over five datasets, rest on at least one incorrectly predicted concept, against 9.6\% for our method ($\mathrm{RCA}_{\mathrm{COR}}$ in Tab.~\ref{tab:main_results}). Such trajectories undermine trust even when the final diagnosis is correct: \emph{a correct prediction does not imply a reliable explanation}. Using predictive uncertainty as a proxy for concept error does not by itself resolve this issue: both uncertainty-aware variants of IP~\cite{nahiduzzaman2026uncertainty}, EUAV-IP and IUAV-IP, are less reliable on average than uncertainty-agnostic Generative IP~\cite{chattopadhyay2022interpretable} (Tab.~\ref{tab:main_results}).
To our knowledge, no existing method measures or controls the correctness of the acquired concept sequence.

Reliability also depends on how concepts are selected: the fewer concept predictions an acquisition policy requires to reach a confident diagnosis, the fewer opportunities there are for prediction errors. Information Pursuit (IP)~\cite{chattopadhyay2022interpretable} explicitly selects the concept with the highest conditional mutual information with the label, but estimating this quantity requires costly generative modelling. Variational methods amortise selection through a querier network trained to approximate the information-gain criterion~\cite{chattopadhyayvariational,nahiduzzaman2026uncertainty}, while reinforcement-learning approaches learn acquisition policies directly from task rewards~\cite{shim2018joint}. Unlike IP, these learned policies predict the next concept to acquire rather than explicitly scoring all remaining candidates, introducing an additional source of error: policy-approximation error, on top of any error in the quantities used to guide acquisition.
However, explicit search is only as effective as the objective it optimises. IP’s conditional information gain has no general fixed-budget approximation guarantee and can undervalue complementary findings, such as an exclusive-or pair whose members are uninformative alone. This matters when diagnostic value arises from combinations of findings, as in multi-criterion dermoscopic rules and biomarker-combination models~\cite{soyer2004three, zalaudek2006three, liu2011min, meisner2021combining}.

We propose \textbf{Re}liable \textbf{C}oncept \textbf{A}cquisition with
\textbf{P}runing (ReCAP), which replaces conditional information gain with an
explicit label-discriminating objective. ReCAP treats the vector of CP outputs,
including abstentions on uncertain concepts, as a hidden state and builds an
implicit graph linking cases with different diagnoses, in the spirit of
equivalence-class edge cutting~\cite{golovin2010near}. Querying a concept cuts
the links between cases whose responses differ, and ReCAP selects the concept
expected to cut the largest probability-weighted mass of links. Because it separates differently diagnosed cases rather than identifying the full concept state, it credits complementary concepts such as an exclusive-or pair. The
objective needs only two lightweight predictors, of how the CP will respond to
an unqueried concept and of the diagnosis after each response, so ReCAP avoids
IP's generative modelling and requires no policy network. Abstention is a valid
outcome, so concepts the CP is unlikely to answer are deprioritised. Finally, a
diagnosis-preserving pruning step removes acquired concepts that do not change
the prediction, shortening the explanation without sacrificing accuracy.

\noindent Our main contributions are:
\begin{itemize}\itemsep2pt
\item We formalize explanation reliability as a goal distinct from diagnostic
      accuracy, introduce the \emph{reliable concept acquisition rate} (RCA) to
      measure it, and formulate acquisition over CP-reported concepts, with
      abstention treated as an explicit outcome (Sec.~\ref{sec:problem}).
\item We cast acquisition under imperfect concept prediction as equivalence-class edge cutting~\cite{golovin2010near} over CP-reported states, and show that its uncut edge weight reduces to $p(\mathcal{H})^2G(\mathcal{H})$, so the objective can be computed from two learned heads without enumerating states. Its adaptive monotonicity and submodularity yield a $(1-1/e)$ fixed-budget guarantee, which degrades by at most $2k\epsilon$ under learned estimates with uniform error $\epsilon$ (Secs.~\ref{subsec:surrogate},~\ref{subsec:guarantees}).
\item We show that the remaining diagnostic uncertainty (Gibbs error) is a principled stopping rule, bounding both the diagnostic error and the gap between the explanation’s conclusion and the full image’s (Prop.~\ref{prop:decomp}, Cor.~\ref{cor:sufficiency}).
\item On five medical imaging datasets, ReCAP raises average RCA from 65.36\% to 84.71\% over the strongest baseline and reduces the share of correct diagnoses resting on an incorrect concept from 27.1\% to 9.6\%, while using fewer than half as many concepts and achieving higher diagnostic accuracy. On OAI, where the
      optimal policy is computable, ReCAP attains at least 98.9\% of its value.
\end{itemize}

\section{Related Work}
\label{sec:related}

\paragraph{Concept-Based Models.} CBMs~\cite{koh2020concept} predict labels through explicit concepts, with subsequent work developing richer representations and architectures~\cite{wickramanayake2021comprehensible, patricio2023coherent,sarkar2022framework,rao2024discover}. Other approaches reduce annotation requirements using pretrained vision-language models or priors~\cite{oikarinen2023label,yang2023language, nahiduzzaman2026weakly}, or model concept uncertainty~\cite{kim2023probabilistic,gao2024evidential}. These advances concern concept construction and representation; our focus is selecting among imperfect concept predictions.

\paragraph{Sequential Concept Acquisition.} Interactive CBMs~\cite{chauhan2023interactive} prioritise concepts for human correction using uncertainty, expected improvement, and intervention cost. Related work on Active Feature Acquisition (AFA)
jointly optimises acquisition and prediction: Joint-AFA~\cite{shim2018joint}
trains a classifier alongside a reinforcement-learning agent that
learns feature acquisition and stopping. However, Joint-AFA operates on
features rather than interpretable concepts, does not explicitly
optimise explanation reliability. Information Pursuit (IP)~\cite{chattopadhyay2022interpretable} selects semantic queries through conditional information gain, while V-IP~\cite{chattopadhyayvariational} learns acquisition and prediction without an explicit generative model. Query-generation extensions use vision-language models~\cite{chattopadhyay2024bootstrapping} or learn binary queries from CLIP representations~\cite{kolek2025learning}; these are complementary to acquisition over a fixed vocabulary. Uncertainty-aware variants (UAV-IPs)~\cite{nahiduzzaman2026uncertainty} either mask uncertain concepts (EUAV-IP) or penalise their selection (IUAV-IP). 
Our approach incorporates abstention as an outcome and explicitly evaluates candidates under an objective supporting a fixed-budget approximation guarantee.

\paragraph{Adaptive Submodularity.} Adaptive submodularity~\cite{golovin2011adaptive} formalises conditional diminishing returns and supports approximation guarantees for adaptive greedy selection. Prior work establishes guarantees for noisy Bayesian active learning through edge-cutting objectives~\cite{golovin2010near}, identifies conditions for near-optimal greedy information maximisation~\cite{chen2015sequential}, and develops Gibbs-error-based active learning~\cite{cuong2013active} and adaptively submodular surrogates for value of information~\cite{chen2015submodular}. Related work provides approximation algorithms for decision trees with persistent noisy outcomes~\cite{jia2019optimal}. We follow the surrogate approach, representing imperfect concept acquisition as coordinate revelation of a reproducible predictor-output state without assuming independent concept errors.

\section{Problem Formulation} 
\label{sec:problem} 

For an image $x\in\mathcal{X}$ with label $y\in\mathcal{Y}$, let $c^\star(x)\in\{-1,+1\}^M$ denote the vector of $M$ true binary concepts, with $-1$ and $+1$ indicating absence and presence, respectively. Following~\cite{chattopadhyay2022interpretable}, we assume:
\begin{assumption}[Concept sufficiency~\cite{chattopadhyay2022interpretable}]
\label{ass:sufficiency}
The label distribution depends on the image only through its true concepts:
\begin{equation}
\label{eq:sufficiency}
P(y\mid x)=P(y\mid c^\star(x)).
\end{equation}
\end{assumption}

\paragraph{Concept observation model.} 
At inference, true concepts are unavailable. A fixed CP provides a confidence $\mu_m(x)$ estimating the presence of each concept for a given image $x$, where $m\in\{1,\dots,M\}$. Following UAV-IPs~\cite{nahiduzzaman2026uncertainty}, we compute the associated uncertainty $\sigma_m(x)$ using Monte Carlo dropout~\cite{gal2016dropout}, although other uncertainty estimators may be used.

\paragraph{Instance-specific reliability and abstention.} A signed concept prediction is \emph{reliable} if it equals the true value and \emph{unreliable} otherwise, i.e.: $\hat{c}_m(x)=c^\star_m(x)$. Even a concept with a high average score can fail on a specific image. Since correctness is unobservable at inference, we use input-dependent uncertainty as a proxy and define 
\begin{equation}
\label{eq:admissible}
\hat c_m(x)= \begin{cases} \operatorname{sign}\big(\mu_m(x)-\tfrac12\big), & \sigma_m(x)\leq\kappa,\\ \bot, & \text{otherwise}. \end{cases}
\end{equation}
where $\kappa$ is a fixed threshold and $\bot$ denotes abstention, which makes no concept assertion. Thus,
\begin{equation} 
\label{eq:chat} 
\hat c(x)=[\hat c_1(x),\dots,\hat c_M(x)] \in\mathcal{A}^M, \qquad \mathcal{A}=\{-1,+1,\bot\}. 
\end{equation}

\begin{assumption}[Reproducibility] 
\label{ass:determinism} 
For each image, the observable vector $\hat c(x)$ is fixed: querying concept $m$ always returns the same value $\hat c_m(x)$. 
\end{assumption} 

\paragraph{Sequential acquisition and objective.} 
A policy $\pi$ sequentially queries unacquired concepts or stops and emits a diagnosis. Its decisions depend only on acquired concept outcomes, without access to the image or unqueried confidences and uncertainties. The acquisition trajectory and history are 
\[ 
\tau^\pi(x)=(j_1,\dots,j_{|\tau^\pi(x)|}), \quad \mathcal{H}^\pi_{1:k}(x) =\big[(j_\ell,\hat c_{j_\ell}(x))\big]_{\ell=1}^k. 
\] 
Let $p(y\mid\mathcal{H})$ denote the diagnostic distribution conditioned on the acquired history. We seek a faithful diagnosis with minimal expected query cost~\cite{chattopadhyay2022interpretable}: 
\begin{equation} 
\label{eq:base} 
\begin{aligned} 
\min_\pi\quad &\mathbb{E}_x[|\tau^\pi(x)|]\\ \text{s.t.}\quad &\mathbb{E}_x\Big[ d\big( P(y\mid x), p(y\mid\mathcal{H}^\pi_{1:|\tau^\pi(x)|}(x)) \big) \Big]\leq\delta, 
\end{aligned} 
\end{equation} where $d$ is a divergence and $\delta>0$ is the fidelity tolerance. 

Even acquiring every concept need not recover $P(y\mid x)$, because the CP may discard label-relevant information\footnote{By Assumption~\ref{ass:sufficiency} and data processing, $I(y;\hat c)\leq I(y;c^\star)=I(y;x)$, with equality only when $\hat c$ preserves all label-relevant information.}. The best fidelity achievable from $\hat c$ is
\begin{equation} 
\label{eq:delta_min} 
\delta_{\min}\triangleq \inf_q\mathbb{E}_x\Big[ d\big(P(y\mid x),q(y\mid\hat c(x))\big) \Big], 
\end{equation} 
so Equation~\eqref{eq:base} is infeasible for $\delta<\delta_{\min}$, whatever the policy.

\section{Method}
\label{sec:method}
Our method is, in essence, an implicit graph cutting algorithm. Each node is a CP output vector paired with a diagnosis it co-occurs with in the training data, and nodes with different diagnoses are joined by edges. Each answer on the test image rules out the nodes that disagree with it, cutting their edges, and ReCAP adaptively selects queries until no edges remain among the surviving nodes, which then share a common diagnosis. We formalise the states, then the objective and its properties, and finally its estimation with learned heads and the resulting guarantees.

\paragraph{CP outputs as a hidden state.} Under Assumption~\ref{ass:determinism}, $\hat c(x)\in\mathcal{A}^M$ is fixed given $x$, so we treat it as a latent state with joint prior $p(\hat c,y)$; these states are the nodes above. 
Acquisition is thus adaptive stochastic optimisation~\cite{golovin2011adaptive}: each query reveals one coordinate of the state and rules out the states that disagree with it. After a history $\mathcal{H}=\{(j_\ell,a_\ell)\}_{\ell=1}^k$, the remaining states form the \emph{version space}
\begin{equation}
\label{eq:version_space}
V(\mathcal{H})\triangleq \big\{v\in\mathcal{A}^M: v_{j_\ell}=a_\ell,\ \ell=1,\dots,k\big\},
\end{equation}
with prior mass $p(\mathcal{H})=\Pr[\hat c\in V(\mathcal{H})]$. Each observation restricts this set, so version spaces are nested.

\begin{figure*}[!t]
\centering
\definecolor{melfill}{HTML}{FAECE7}
\definecolor{melline}{HTML}{993C1D}
\definecolor{nevfill}{HTML}{E1F5EE}
\definecolor{nevline}{HTML}{0F6E56}
\resizebox{.8\textwidth}{!}{%
\begin{tikzpicture}[
  every node/.style={transform shape},
  st/.style={circle, draw, line width=0.4pt, minimum size=5.5mm,
             inner sep=0pt, font=\scriptsize},
  mel/.style={st, fill=melfill, draw=melline, text=melline},
  nev/.style={st, fill=nevfill, draw=nevline, text=nevline},
  ed/.style={draw=melline, opacity=0.75},
  wt/.style={font=\scriptsize, inner sep=1.2pt},
  vec/.style={font=\scriptsize, inner sep=1.2pt},
  hd/.style={font=\small},
  gn/.style={font=\small}]

\node[hd] at (2.3,3.5) {before any query};
\node[mel] (A) at (0.4,2.5) {A};
\node[mel] (B) at (0.4,0.7) {B};
\node[nev] (C) at (4.2,2.5) {C};
\node[nev] (D) at (4.2,0.7) {D};
\node[vec, above] at (A.north) {$(\!+\!,\!+\!,\!-\!)$, $0.3$};
\node[vec, below] at (B.south) {$(\!+\!,\!-\!,\!-\!)$, $0.2$};
\node[vec, above] at (C.north) {$(\!-\!,\!+\!,\!-\!)$, $0.3$};
\node[vec, below] at (D.south) {$(\!-\!,\!-\!,\!+\!)$, $0.2$};
\draw[ed, line width=1.6pt] (A) -- node[wt, above]{$0.18$} (C);
\draw[ed, line width=0.8pt] (B) -- node[wt, below]{$0.08$} (D);
\draw[ed, line width=1.1pt] (A) -- node[wt, pos=0.27, above, sloped]{$0.12$} (D);
\draw[ed, line width=1.1pt] (B) -- node[wt, pos=0.27, above, sloped]{$0.12$} (C);
\node[gn] at (2.3,-0.5) {$R(\varnothing)=0.5$};

\draw[gray, line width=0.3pt] (5.5,-0.7) -- (5.5,3.7);

\node[hd] at (7.3,3.5) {if concept $1$ queried};
\node[wt] at (6.05,2.4) {$a\!=\!+\!1$};
\node[wt] at (6.05,1.0) {$a\!=\!-\!1$};
\node[mel] (A1) at (6.8,2.4) {A};
\node[mel] (B1) at (7.8,2.4) {B};
\node[nev] (C1) at (6.8,1.0) {C};
\node[nev] (D1) at (7.8,1.0) {D};
\node[wt] at (7.3,1.85) {$R\!=\!0$, $f\!=\!0.5$};
\node[wt] at (7.3,0.45) {$R\!=\!0$, $f\!=\!0.5$};
\node[gn] at (7.3,-0.5) {$\mathbb{E}[R]=0$, \ $\Delta=0.5$};

\draw[gray, line width=0.3pt] (9.2,-0.7) -- (9.2,3.7);

\node[hd] at (11.2,3.5) {if concept $2$ queried};
\node[wt] at (9.65,2.4) {$a\!=\!+\!1$};
\node[wt] at (9.65,1.0) {$a\!=\!-\!1$};
\node[mel] (A2) at (10.4,2.4) {A};
\node[nev] (C2) at (12.0,2.4) {C};
\draw[ed, line width=1.6pt] (A2) -- node[wt, above]{$0.18$} (C2);
\node[mel] (B2) at (10.4,1.0) {B};
\node[nev] (D2) at (12.0,1.0) {D};
\draw[ed, line width=0.8pt] (B2) -- node[wt, above]{$0.08$} (D2);
\node[wt] at (11.2,1.85) {$R\!=\!0.18$, $f\!=\!0.32$};
\node[wt] at (11.2,0.45) {$R\!=\!0.08$, $f\!=\!0.42$};
\node[gn] at (11.2,-0.5) {$\mathbb{E}[R]=0.14$, \ $\Delta=0.36$};

\draw[gray, line width=0.3pt] (13.4,-0.7) -- (13.4,3.7);

\node[hd] at (15.35,3.5) {if concept $3$ queried};
\node[wt] at (13.95,2.4) {$a\!=\!+\!1$};
\node[wt] at (13.95,1.0) {$a\!=\!-\!1$};
\node[nev] (D3) at (15.4,2.4) {D};
\node[mel] (A3) at (14.6,1.4) {A};
\node[mel] (B3) at (14.6,0.6) {B};
\node[nev] (C3) at (16.1,1.0) {C};
\draw[ed, line width=1.6pt] (A3) -- node[wt, pos=0.5, above, sloped]{$0.18$} (C3);
\draw[ed, line width=1.1pt] (B3) -- node[wt, pos=0.5, below, sloped]{$0.12$} (C3);
\node[wt] at (15.35,1.9) {$R\!=\!0$, $f\!=\!0.5$};
\node[wt] at (15.35,0.05) {$R\!=\!0.30$, $f\!=\!0.20$};
\node[gn] at (15.35,-0.5) {$\mathbb{E}[R]=0.24$, \ $\Delta=0.26$};
\end{tikzpicture}%
}
\caption{Residual discriminating mass for four states. Each node shows a state's concept values $(\hat c_1,\hat c_2,\hat c_3)$ and prior probability, coloured by diagnosis; edges join states with different diagnoses, with thickness and numbers giving their weights (Eq.~\eqref{eq:edge_weight}). The right panels show the branches that would result from querying each concept, with $a$ the observed value, annotated with residual mass $R$ and resolved mass $f$; each panel also reports the expected residual $\mathbb{E}[R]$ and marginal gain $\Delta$. Concept~1 yields label-pure branches and cuts every edge, so Eq.~\eqref{eq:greedy_rule} selects it even though each branch still contains two states.}
\label{fig:edgecut}
\end{figure*}
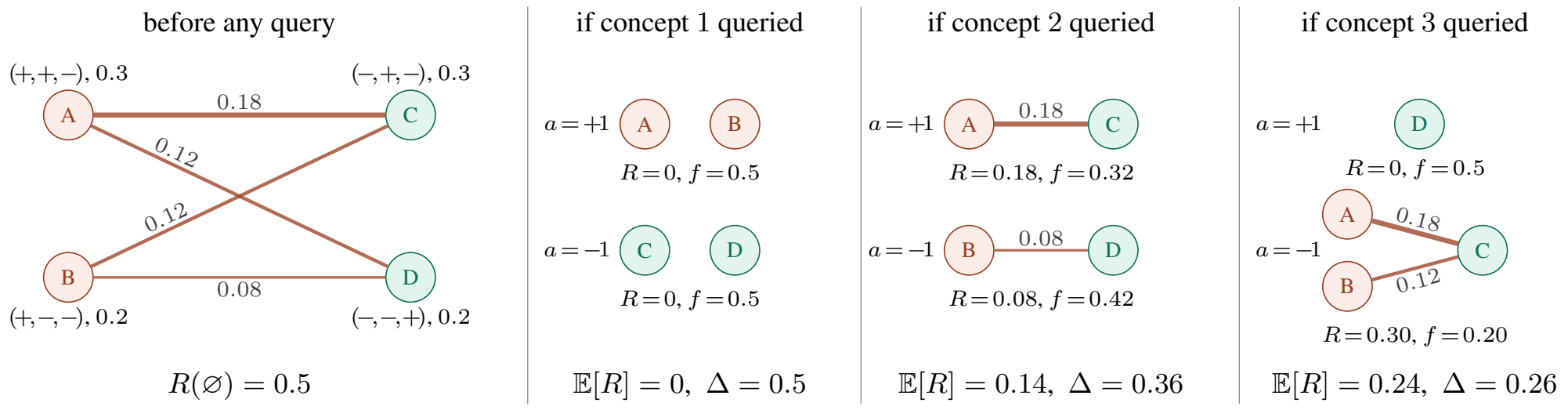

\subsection{A Label-Discriminating Objective} \label{subsec:surrogate} 
As our goal is to separate states by class, our acquisition history (or sequence of cuts) needs only to distinguish cases with different labels, rather than identify the full state. Following equivalence-class edge cutting~\cite{golovin2010near}, with CP-reported states as hypotheses and diagnoses as classes, we construct a graph on the support of $p(\hat c,y)$, connecting differently labelled pairs with weight 
\begin{equation} 
\label{eq:edge_weight} 
w\big(\{(v,y),(v',y')\}\big) =2\,p(v,y)\,p(v',y')\,\mathds{1}[y\neq y']. 
\end{equation} 
Querying a coordinate on which two states differ distinguishes them, cutting their edge. Figure~\ref{fig:edgecut} illustrates this process. 

For a label distribution $b$, define its Gibbs error as $G(b)=1-\sum_y b_y^2$, and write $G(\mathcal{H})=G(p(\cdot\mid\mathcal{H}))$.
The remaining edge mass is 
\begin{align} R(\mathcal{H}) &\triangleq \sum_{\substack{v,v'\in V(\mathcal{H})\\y\neq y'}} p(v,y)\,p(v',y') \nonumber\\ &=p(\mathcal{H})^2 \left[1-\sum_y p(y\mid\mathcal{H})^2\right] =p(\mathcal{H})^2G(\mathcal{H}). 
\label{eq:residual} 
\end{align} 
The equality follows from $\sum_{v\in V(\mathcal{H})}p(v,y) =p(\mathcal{H})p(y\mid\mathcal{H})$. 
We call $R$ the \emph{residual discriminating mass} and maximise the resolved mass, \begin{equation} \label{eq:surrogate} f(\mathcal{H})\triangleq R(\varnothing)-R(\mathcal{H}). \end{equation} Here, $G(\mathcal{H})$ measures residual label uncertainty, while $p(\mathcal{H})^2$ weights it by the probability that two independent states are both consistent with the history.

\paragraph{Diminishing Returns and Fidelity}
ReCAP maximises $f$ greedily, which is near-optimal when $f$ is adaptively monotone and adaptively submodular~\cite{golovin2011adaptive}. 
For an unacquired concept $m$, define the expected marginal gain
\begin{equation}
\label{eq:gain_def}
\Delta(m\mid\mathcal{H})\triangleq \mathbb{E}_{\hat c_m\mid\mathcal{H}} \big[f(\mathcal{H}\cup\{(m,\hat c_m)\})-f(\mathcal{H})\big].
\end{equation}
\begin{proposition}[Diminishing returns; adapted from~\cite{golovin2010near}]
\label{prop:adasub}
Under Assumption~\ref{ass:determinism}, for every positive-probability history $\mathcal{H}$ and unacquired concept $m$, $\Delta(m\mid\mathcal{H})\geq0$ (adaptive monotonicity), and $\Delta(m\mid\mathcal{H})\geq\Delta(m\mid\mathcal{H}')$ for every positive-probability extension $\mathcal{H}'\supseteq\mathcal{H}$ in which $m$ is unacquired (adaptive submodularity).
\end{proposition}
\noindent Proofs of all results in this section are in the supplementary material (Secs.~\ref{appendix:proofs}).

A guarantee on $f$ is useful only if its label-uncertainty term $G(\mathcal{H})$ in Equation~\eqref{eq:residual} is meaningful for diagnosis. We therefore relate Gibbs error to diagnostic risk and to fidelity, measured by the Brier divergence $d_{\mathrm{B}}(b,b')=\sum_y(b_y-b'_y)^2$, which yields an exact decomposition and, unlike KL divergence, remains finite when a posterior excludes a label.
\begin{proposition}[cf.~\cite{cover1967nearest}]
\label{prop:sandwich}
For any distribution $b$ over $\mathcal{Y}$, the Gibbs error $G(b)=1-\sum_y b_y^2$ and Bayes error $B(b)=1-\max_y b_y$ satisfy $B(b)\leq G(b)\leq B(b)\big(2-\tfrac{|\mathcal{Y}|}{|\mathcal{Y}|-1}B(b)\big)$.
\end{proposition}
Thus, controlling Gibbs error controls MAP classification risk. Fidelity instead concerns the gap between the posterior given the acquired concepts and the full-image posterior; the following Brier-score decomposition relates the two exactly.
\begin{proposition}[Fidelity decomposition]
\label{prop:decomp}
Let $\pi$ be a deterministic policy whose query and stopping decisions depend only on acquired outcomes, let $\mathcal{H}$ be its terminal history, let $p(y\mid\mathcal{H})$ be the population Bayes posterior, and write $G_x=G(P(\cdot\mid x))$. Then
\begin{equation}
\label{eq:fidelity_identity}
\mathbb{E}_x\big[ d_{\mathrm{B}}(P(y\mid x),p(y\mid\mathcal{H})) \big] =\mathbb{E}[G(\mathcal{H})]-\mathbb{E}[G_x].
\end{equation}
\end{proposition}
\begin{corollary}
\label{cor:sufficiency}
Let $\pi$ be as in Proposition~\ref{prop:decomp}, and suppose its terminal history satisfies $\mathbb{E}[G(\mathcal{H})]\leq\delta'$ and it emits the MAP diagnosis $\hat y(\mathcal{H})\in\arg\max_y p(y\mid\mathcal{H})$, with ties broken by a fixed rule. Then its misclassification probability is at most $\delta'$, and it satisfies the fidelity constraint of Equation~\eqref{eq:base} with $d=d_{\mathrm{B}}$ and $\delta=\delta'$.
\end{corollary}
Together, these results justify Gibbs error as a stopping criterion for both diagnostic accuracy and fidelity. Equation~\eqref{eq:fidelity_identity} identifies the gap between them as the intrinsic task ambiguity $\mathbb{E}[G_x]$, distinct from the CP-induced floor $\delta_{\min}$ (Prop.~\ref{prop:floor} in supplementary). These identities assume the population Bayes posterior, which finite-data classifiers only approximate.

\subsection{Selection Rule} 
\label{subsec:algorithm} 
Substituting Equation~\eqref{eq:residual} into Equation~\eqref{eq:gain_def} gives 
\begin{align} 
\Delta&(m\mid\mathcal{H}) =R(\mathcal{H}) -\sum_{a\in\mathcal{A}}p(a\mid m,\mathcal{H}) R(\mathcal{H}\cup\{(m,a)\}) \nonumber\\ &=p(\mathcal{H})^2 \Big[ G(\mathcal{H}) -\sum_{a\in\mathcal{A}}p(a\mid m,\mathcal{H})^3 G(\mathcal{H}\cup\{(m,a)\}) \Big], 
\label{eq:gain_expanded} 
\end{align} 
where $p(a\mid m,\mathcal{H})=\Pr(\hat c_m=a\mid\mathcal{H})$. The cubic weight combines the answer probability with the squared branch mass in $R$. Let $\mathcal{H}=\mathcal{H}_{1:k}$ and define 
\(\mathcal{C}_k\triangleq \{1,\dots,M\}\setminus\mathrm{dom}(\mathcal{H}_{1:k}). \)
Since $p(\mathcal{H})^2$ and $G(\mathcal{H})$ are common to all candidates, maximising marginal gain is equivalent to 
\begin{equation} 
\label{eq:greedy_rule} j_{k+1}\in\arg\min_{m\in\mathcal{C}_k} \sum_{a\in\mathcal{A}}p(a\mid m,\mathcal{H})^3 G(\mathcal{H}\cup\{(m,a)\}). 
\end{equation}  

The rule requires an \emph{answer predictive} $p_\psi(a\mid m,\mathcal{H})$ and a \emph{label posterior} $p_\eta(y\mid\mathcal{H})$. The former predicts all candidates as an $M\times|\mathcal{A}|$ tensor; the latter is the diagnostic classifier used in V-IP-style~\cite{chattopadhyayvariational} architectures. Neither head explicitly estimates $p(\hat c,y)$ or mutual information. Define 
\(
G_\eta(\mathcal{H})\triangleq 1-\sum_y p_\eta(y\mid\mathcal{H})^2 
\)
and the learned score 
\begin{equation}
\label{equ:learned_score}
s_\eta(m\mid\mathcal{H}) =\sum_{a\in\mathcal{A}}p_\psi(a\mid m,\mathcal{H})^3 G_\eta(\mathcal{H}\cup\{(m,a)\}).
\end{equation} 
Ranking by $s_\eta$ never requires $p(\mathcal{H})$; the analysis of learned
gains (Sec.~\ref{subsec:guarantees}) uses the corresponding absolute gain
$\widehat\Delta(m\mid\mathcal{H})=p(\mathcal{H})^2\big(G_\eta(\mathcal{H})-s_\eta(m\mid\mathcal{H})\big)$.


\paragraph{Training the predictive heads.} We initialise $p_\eta$ on complete histories, jointly train both heads on randomly sampled partial histories, and fine-tune on greedy acquisition trajectories. Both use cross-entropy: $p_\eta$ predicts labels, while $p_\psi$ predicts CP outputs without concept annotations. No separate policy network is trained. Losses and training details appear in Sec.~\ref{appendix:training}.

\paragraph{Pruning the delivered explanation.} A query may reduce discriminating mass without contributing to the terminal diagnosis. We therefore prune the terminal history sequentially, removing a concept $m$ only when \begin{equation} \label{eq:inert_approx} \begin{aligned} \big\| p_\eta(\cdot\mid\mathcal{H}\setminus\{m\}) -p_\eta(\cdot\mid\mathcal{H}) \big\|_1&\leq\xi,\\ \hat y(\mathcal{H}\setminus\{m\})&=\hat y(\mathcal{H}), \end{aligned} \end{equation} where $\mathcal{H}\setminus\{m\}$ removes the observation indexed by $m$. 
Further justification appears in Sec.~\ref{appendix:pruning}.
Acquisition stops when the tolerance is met, the cap is reached, or candidates are exhausted. 
The complete acquisition and pruning procedure is provided in
Algorithm~\ref{alg:greedy} in the supplementary material~\ref{appendix:greedy_algorithm}.

\subsection{Guarantees} 
\label{subsec:guarantees} 
Proposition~\ref{prop:adasub} permits the standard adaptive greedy guarantee~\cite{golovin2011adaptive}, comparing objective value under a fixed query budget. 
\begin{theorem}[Budgeted guarantee] 
\label{thm:budget} 
Under Assumption~\ref{ass:determinism}, the population greedy policy run for $k$ acquisitions satisfies 
\begin{equation} 
\label{eq:budget_bound} 
f_{\mathrm{avg}}(\pi_{\mathrm{gr}}^{[k]}) \geq \left(1-\frac{1}{e}\right) \max_{\pi:\,|\tau^\pi|\leq k}f_{\mathrm{avg}}(\pi), 
\end{equation} 
where $f_{\mathrm{avg}}(\pi)=\mathbb{E}[f(\mathcal{H}^\pi)]$ and the comparator budget holds almost surely. 
\end{theorem} 

For learned scores, a uniform absolute marginal-gain error of at most $\epsilon$ weakens this bound by $2k\epsilon$ (Proposition~\ref{prop:approx}). Both results concern fixed-horizon acquisition before pruning and compare policies observing $\hat c$, not $c^\star$. 
Corollary~\ref{cor:sufficiency} provides the separate fidelity guarantee under the population Bayes posterior. Proofs of all results, together with full statements and estimation caveats, are in the supplementary material (Sec.~\ref{appendix:guarantees}).

\section{Experiments}

\subsection{Synthetic Analysis}

To isolate the effect of the selection objective, we simulate acquisition in which the relationship between concepts and diagnosis, image-level posterior, and optimal policy are all known (synthetic data explained in Appendix Sec.~\ref{appendix:synthetic}). All methods use the same probabilities, either counted from the simulated cases or learned as in Sec.~\ref{sec:method}, and differ only in the concept-selection rule, so ReCAP’s objective is compared directly with information gain.
Each case has eight binary
concepts that stylise patterns seen in medical
diagnosis~\cite{soyer2004three,meisner2021combining}: a pair of findings that
is diagnostic only in combination; three rare findings that are decisive when present; and three weak
cues present in 15\% of cases. The predictor misreads 5\% of every concept, with
errors correlated within each image. 

\begin{figure*}[t]
\centering
\includegraphics[width=\textwidth]{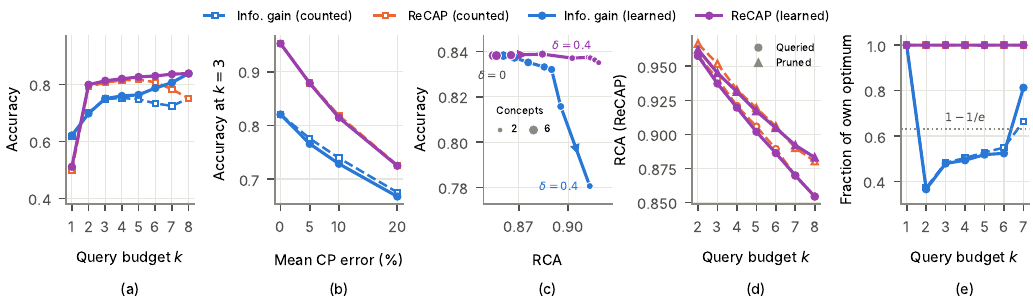}
\caption{Synthetic analysis (eight concepts: an exclusive-or pair,
three rare decisive concepts, three weak cues). Solid: learned heads; dashed: counted probabilities.
(a) Accuracy against budget with 5\% predictor
errors and abstention.
(b) Accuracy at $k=3$ against mean predictor error, without abstention.
(c) Accuracy against RCA under Gibbs stopping with abstention, traced by
$\delta\in[0,0.4]$; arrows: increasing $\delta$; marker area: concepts
asserted.
(d) ReCAP's RCA before (circles) and after (triangles) pruning, $\xi=0.05$,
with abstention.
(e) Greedy value of each rule's own objective as a fraction of that
objective's exact optimum, without abstention (ReCAP: $f$; info.\ gain:
terminal entropy); dotted: $1-1/e$.}
\label{fig:synthetic}
\end{figure*}

\noindent \emph{Complementary evidence.} Information gain favors individually
informative rare concepts and plateaus with 0.70--0.81 accuracy for
$k=2$--$7$. ReCAP instead acquires the complementary pair first and reaches 0.799 at $k=2$
and 0.813--0.836 from $k=3$, close to the 0.838 obtained by querying every concept
(Fig.~\ref{fig:synthetic}a). This advantage persists, though it narrows, as predictor error grows (Fig.~\ref{fig:synthetic}b). The advantage does not hold
at $k=1$ (0.511 vs.\ 0.621), since either member of the pair is uninformative
alone. Counted
probabilities match the learned heads to within 0.014 up to $k=4$ but fall to 0.752 with every concept
queried: with abstention, each concept has three outcomes, so long histories are rare and 15\% of
complete test histories never occur in training. The learned heads share parameters across histories
and do not degrade.

\noindent \emph{Reliability and explanation length.} At a fixed budget, all methods attain the same RCA to within 0.8 points, so reliability differs only through explanation length\footnote{The simulated predictor misreads every concept at the same 5\% rate, so no selection rule can find more reliable concepts. On real data, CP errors vary across concepts, so selection can also avoid unreliable concepts (Sec.~\ref{sec:supp_fixed_budget}).}. Under Gibbs stopping ($\delta=0.30$, with abstention), ReCAP asserts 2.34 concepts against 4.20 for information gain, with slightly higher accuracy (0.837 vs.\ 0.832) and higher RCA (0.913 vs.\ 0.890). As $\delta$ grows, ReCAP gains RCA at almost constant accuracy, whereas information gain must give up accuracy (Fig.~\ref{fig:synthetic}c). Without abstention the gap widens (2.00 vs.\ 5.15 concepts; RCA 0.910 vs.\ 0.823), although ReCAP’s per-assertion error is not lower (4.8\% vs.\ 4.2\%), so the gain comes from shorter explanations alone. Pruning further shortens explanations at unchanged accuracy ($k=3$: 2.40 to 1.92 concepts, RCA 0.937 to 0.945; Fig.~\ref{fig:synthetic}d).

\noindent \emph{Guarantees.} Exact dynamic programming shows that greedy ReCAP attains
the optimal expected objective at every budget, also when it selects with learned heads, whereas
information gain attains only 37--66\% of its own optimum for $k=2$--$7$ (37--81\% with learned heads; Fig.~\ref{fig:synthetic}e). Under Gibbs stopping, the
diagnostic error remains below $\mathbb{E}[G(\mathcal{H})]$, and the Brier
discrepancy to the image posterior agrees with
Proposition~\ref{prop:decomp} to within 0.008.
 
The conclusions largely hold when the interaction
strength and the prevalence of the weak cues and rare concepts are varied
(Sec.~\ref{appendix:synthetic}, Fig.~\ref{fig:synthetic_regimes}). The
advantage narrows as weak cues approach balance or rare concepts approach a
third of cases, and reverses when the interaction is weak or absent, where
ReCAP's preference for balanced splits still selects a pair member first.

\subsection{Real-Data Experimental Setup}
\paragraph{Datasets.}
We evaluate our method on five public medical imaging datasets spanning five
imaging modalities, with concept spaces ranging from 4 to 29 concepts. In
increasing order of concept-space size, these include retinal fundus
photography (DDR~\cite{bian2024diabetic}; 1{,}500 images, 2 classes,
4 lesion concepts), breast ultrasound (BrEaST~\cite{pawlowska2024curated};
256 images, 2 classes, 7 BI-RADS descriptors), dermoscopy
(PH2~\cite{mendoncca2013ph}; 200 images, 2 classes, 8 concepts), knee X-ray
(OAI~\cite{nevitt2006osteoarthritis}; 4{,}506 patients, 4
Kellgren--Lawrence grades, 10 clinical concepts), and microscopy
(WBCatt~\cite{tsutsui2023wbcatt}; 10{,}198 blood smear images, 5 cell types,
29 morphological concepts). These five datasets cover five distinct
modalities and diverse dataset and concept-space sizes. Following
UAV-IPs~\cite{nahiduzzaman2026uncertainty}, OAI is formulated as a four-class classification task.
Dataset splits and complete concept lists are provided in the supplementary
material Sec~\ref{appendix:datasets}.


\paragraph{Evaluation metrics.}
We propose the Reliable Concept Acquisition rate (RCA), the percentage of test samples whose explanation contains no incorrectly predicted concept. With $W_i=\sum_{j\in\tau_i}\mathbb{I}[\hat c_{ij}\neq c_{ij}^\ast]$ the number of such concepts for sample $i$,

\begin{equation}
\label{eq:rca_metrics}
\begin{aligned}
\mathrm{RCA}_{\mathrm{ALL}}
&=\frac{100}{N}\sum_{i=1}^{N}\mathbb{I}[W_i=0],\\
\mathrm{RCA}_{\mathrm{COR}}
&=\frac{100}{N_{\mathrm{cor}}}
\sum_{i:\hat y_i=y_i}\mathbb{I}[W_i=0].
\end{aligned}
\end{equation}

where $N_{\mathrm{COR}}$ is the number of correctly diagnosed samples, so $\mathrm{RCA}_{\mathrm{COR}}$ measures reliability given a correct diagnosis. We also report Macro-F1 (MF1), accuracy (ACC), and the average number of concepts in the explanation ($Q_{\mathrm{avg}}$).



\paragraph{Implementation details.}
All models are implemented in PyTorch and trained on an NVIDIA Quadro RTX 8000 GPU (48 GB). The CP uses a ResNet-101 backbone for PH2 and WBCatt and Inception-v3 for DDR, BrEaST and OAI, with uncertainty from Monte Carlo dropout. Hyperparameters are selected on the validation set; architectures, training settings and concept-level reliability statistics are in Supplementary material Secs.~\ref{appendix:implementation} and~\ref{sec:supp_selective_reliability}. 

\begin{table*}[t]
\centering
\caption{
{Comparison of concept acquisition methods across five medical imaging datasets.}
Seq.: sequential acquisition; Amort.: concepts selected by a trained policy or querier network rather than by explicitly scoring every candidate; Unc.: uses concept-prediction uncertainty; Guar.: fixed-budget approximation guarantee for the selection rule.
Dataset-specific columns report Reliable Concept Acquisition Rate
over all test samples ($\mathrm{RCA}_{\mathrm{ALL}}$).
The final five columns report unweighted cross-dataset averages
of $\mathrm{RCA}_{\mathrm{ALL}}$, $\mathrm{RCA}_{\mathrm{COR}}$ (restricted to correctly classified samples),
Macro-F1 (MF1), accuracy (ACC), and $Q_{\mathrm{avg}}$.
GISS-U is a ground-truth-informed diagnostic reference that uses the
true diagnosis to select candidate histories without correcting
concept predictions; it is excluded from method ranking.
Best and second-best among deployable methods
are shown in \textbf{bold} and \underline{underlined}, respectively.
}
\label{tab:main_results}
\scriptsize
\setlength{\tabcolsep}{3pt}
\renewcommand{\arraystretch}{1.05}
\begin{tabular}{@{}lcccc|ccccc|ccccc@{}}
\toprule
&&&&& \multicolumn{5}{c|}{\textbf{Trajectory Reliability ($\mathrm{RCA}_{\mathrm{ALL}}$ $\uparrow$)}} &
\multicolumn{5}{c}{\textbf{Cross-Dataset Average}} \\
\cmidrule(lr){6-10}
\cmidrule(lr){11-15}
\textbf{Method} & \textbf{Seq.} & \textbf{Amort.} & \textbf{Unc.} & \textbf{Guar.}
& \textbf{DDR} & \textbf{BrEaST} & \textbf{PH2}
& \textbf{OAI} & \textbf{WBCatt}
& \textbf{$\mathrm{RCA}_{\mathrm{ALL}}$}$\uparrow$
& \textbf{$\mathrm{RCA}_{\mathrm{COR}}$}$\uparrow$
& \textbf{MF1}$\uparrow$
& \textbf{ACC}$\uparrow$
& $\mathbf{Q}_{\mathrm{avg}}\downarrow$ \\
\midrule
CBM~\cite{koh2020concept} & \xmark & -- & \xmark & --
& 56.44 & 25.49 & 35.00 & 65.79 & 67.05
& 49.95 & 53.64 & \underline{80.26} & 84.49 & 11.60 \\
ProbCBM~\cite{kim2023probabilistic} & \xmark & -- & \cmark & --
& 57.67 & 25.49 & 40.00 & 46.84 & 47.67
& 43.53 & 47.15 & 78.59 & 83.13 & 11.60 \\
Evi-CEM~\cite{gao2024evidential} & \xmark & -- & \cmark & --
& 65.33 & 29.51 & 27.50 & 35.17 & 70.25
& 45.55 & 51.22 & 77.96 & 83.47 & 11.60 \\
\arrayrulecolor{gray!50}\specialrule{0.3pt}{1pt}{1pt}\arrayrulecolor{black}
Joint-AFA~\cite{shim2018joint} & \cmark & \cmark & \xmark & \xmark
& \underline{77.23} & \underline{72.23} & 32.50 & 65.63 & 76.80
& 64.88 & 67.40 & 77.44 & 83.09 & 5.91 \\
Generative IP~\cite{chattopadhyay2022interpretable} & \cmark & \xmark & \xmark & \xmark
& 61.39 & 47.06 & \underline{55.83} & \underline{70.02} & \underline{92.51}
& \underline{65.36} & \underline{72.93} & 79.85 & 83.81 & \underline{4.71} \\
V-IP~\cite{chattopadhyayvariational} & \cmark & \cmark & \xmark & \xmark
& 56.44 & 25.49 & 37.50 & 66.53 & 64.21
& 50.03 & 58.85 & 77.46 & 82.81 & 7.16 \\
EUAV-IP~\cite{nahiduzzaman2026uncertainty} & \cmark & \cmark & \cmark & \xmark
& 75.74 & 37.25 & 55.00 & 69.90 & 82.28
& 64.03 & 69.64 & 79.42 & 84.23 & 5.50 \\
IUAV-IP~\cite{nahiduzzaman2026uncertainty} & \cmark & \cmark & \cmark & \xmark
& 59.41 & 43.14 & 35.00 & 68.01 & 83.31
& 57.77 & 67.00 & 79.57 & \underline{84.58} & 6.01 \\
ReCAP (Ours) & \cmark & \xmark & \cmark & \cmark
& \textbf{83.20} & \textbf{73.33} & \textbf{80.83} & \textbf{90.42} & \textbf{95.77}
& \textbf{84.71} & \textbf{90.41} & \textbf{81.40} & \textbf{85.83} & \textbf{2.02} \\
\midrule
\rowcolor{gray!15}
\multicolumn{5}{l|}{\textit{GISS-U (Reference)}}
& 96.53 & 94.12 & 79.17 & 98.39 & 93.27
& 92.30 & 92.57 & 93.36 & 96.07 & 1.49 \\
\bottomrule
\end{tabular}
\end{table*}

\paragraph{Baselines and Evaluation Protocol.}
We select baselines covering concept uncertainty and sequential acquisition\footnote{Automatic concept annotations, concept discovery and query-learning methods are complementary: they construct vocabularies usable upstream, whereas we study acquisition of a given set of imperfect concept predictions.}. Non-sequential baselines include CBM~\cite{koh2020concept}, ProbCBM~\cite{kim2023probabilistic} and
Evi-CEM~\cite{gao2024evidential}; the latter two model concept uncertainty. Sequential baselines include genrative IP~\cite{chattopadhyay2022interpretable}, and V-IP~\cite{chattopadhyayvariational}, the standard information-pursuit baselines; EUAV-IP~\cite{nahiduzzaman2026uncertainty}, and IUAV-IP~\cite{nahiduzzaman2026uncertainty}, which exclude and incorporate concept uncertainty into acquisition respectively; Joint-AFA~\cite{shim2018joint}, a reinforcement-learning-based acquisition method adapted to query human-interpretable concepts rather than raw features.
All sequential methods use the same pretrained CP and binary concept
representation, so they differ only in their acquisition and stopping policies.
Non-sequential baselines use the full concept set for diagnosis; for
matched-budget RCA evaluation, their top-$K$ concepts are selected by absolute
classifier weight, with $K$ matched to our method's average query count. No
method receives human intervention at test time.

\subsection{Comparison with Existing Methods}
\paragraph{Overall comparison.}
Table~\ref{tab:main_results} compares ReCAP with non-sequential
and sequential concept-based methods across five medical imaging
datasets. ReCAP achieves the highest average RCA among deployable
methods, reaching 84.71\%. 
It also achieves the highest average diagnostic performance
(MF1: 81.40\%; ACC: 85.83\%), while retaining only 2.02 concepts
on average. Non-sequential methods, which use the complete concept
set, exhibit substantially lower concept reliability.
These results demonstrate that ReCAP produces more reliable
concept-based explanations while maintaining strong diagnostic
performance. Complete per-dataset results over three random seeds
are reported in Supplementary Table~\ref{tab:main_full_results}.


On $\mathrm{RCA}_{\mathrm{COR}}$, which conditions on a correct diagnosis, ReCAP (90.41\%) comes within 2.2 points of GISS-U (92.57\%), a reference that uses the same CP outputs and uncertainty threshold but selects concept histories using the true diagnosis. ReCAP even exceeds GISS-U’s $\mathrm{RCA}_{\mathrm{ALL}}$ on PH2 (80.83\% versus 79.17\%) and WBCatt (95.77\% versus 93.27\%). This is because GISS-U prefers whichever history yields the correct diagnosis, so when a mispredicted concept happens to push the classifier towards the true class, GISS-U (more details in Supplementary Sec.~\ref{sec:supp_giss}) favours it; ReCAP never sees the label and has no such incentive.

To investigate whether higher RCA simply results from shorter
acquisition trajectories, we further compare methods under fixed
acquisition budgets on WBCatt. ReCAP maintains RCA above 90\%
across budgets of 2--10 concepts, whereas competing methods
generally exhibit declining reliability as the budget increases.
ReCAP also maintains strong diagnostic Macro-F1 at larger budgets,
indicating that its reliability advantage is not solely attributable
to shorter trajectories but also reflects more reliable concept
selection. Detailed results are provided in Supplementary
Section~\ref{sec:supp_fixed_budget}. 


\begin{table}[t]
\centering
\footnotesize
\caption{
{Ablation of ReCAP components.}
Effects of trajectory fine-tuning (TF) and
terminal pruning (TP) on DDR and OAI.
Results are means over three seeds.
}
\label{tab:recap_ablation}
\setlength{\tabcolsep}{3.5pt}
\renewcommand{\arraystretch}{0.95}
\begin{tabular}{lcc|cccc}
\toprule
\textbf{Data}
& \textbf{TF}
& \textbf{TP}
& $\mathbf{\mathrm{RCA}_{\mathrm{ALL}}}\uparrow$
& \textbf{MF1}$\uparrow$
& \textbf{ACC}$\uparrow$
& $\mathbf{Q}_{avg}\downarrow$ \\
\midrule

\multirow{3}{*}{DDR}
& \xmark & \xmark
& 82.91
& 81.04
& 82.93
& 2.73 \\

& \cmark & \xmark
& 83.20
& 82.04
& 83.47
& 1.82 \\

& \cmark & \cmark
& \textbf{83.20}
& \textbf{82.04}
& \textbf{83.47}
& \textbf{1.68} \\

\midrule

\multirow{3}{*}{OAI}
& \xmark & \xmark
& 86.98
& 57.92
& 73.10
& 5.51 \\

& \cmark & \xmark
& 88.25
& 60.21
& 74.83
& 5.36 \\

& \cmark & \cmark
& \textbf{90.42}
& \textbf{60.21}
& \textbf{74.83}
& \textbf{2.86} \\

\bottomrule
\end{tabular}
\end{table}





\subsection{Ablations}
\label{sec:ablations}
Table~\ref{tab:recap_ablation} ablates trajectory fine-tuning (TF) and terminal pruning (TP) on DDR (4 concepts) and OAI (10 concepts). TF, which adapts the heads to histories visited by the acquisition policy, improves RCA, MF1 and accuracy on both datasets and shortens explanations on DDR (2.73 to 1.82 concepts). TP leaves diagnostic performance unchanged by construction while shortening explanations further, most clearly on OAI, where it nearly halves explanation length (5.36 to 2.86 concepts) and raises RCA from 88.25\% to 90.42\% by removing concepts that do not affect the diagnosis.

\subsection{Empirical Analysis of the Guarantees}
\label{sec:empirical_guarantees}

Theorem~\ref{thm:budget} compares greedy acquisition with the optimal adaptive
policy at the same budget, and Proposition~\ref{prop:approx} bounds the cost of
learned gains. Both gaps are defined relative to the optimal adaptive policy; we therefore evaluate them on OAI ($M=10$, $N=1203$ test images), where this policy can be computed exactly by dynamic programming.

We take $p(\hat c,y)$ to be the empirical distribution of thresholded CP
outputs and labels on the test set, so that $p(\mathcal H)$ and
$p(y\mid\mathcal H)$ are frequencies over the images in $V(\mathcal H)$
(Eq.~\eqref{eq:version_space}) and $f$ follows from
Eqs.~\eqref{eq:residual} and~\eqref{eq:surrogate}. Three policies are scored
under this common $f$ at each budget $k$, each acquiring exactly $k$ concepts: \emph{OPT}, the optimal adaptive
policy; \emph{Greedy}, which maximises the exact gain of
Eq.~\eqref{eq:gain_def}; and \emph{ReCAP}, which ranks candidates by its learned score $s_\eta$ (Sec.~\ref{subsec:algorithm}). OPT and Greedy use the evaluation-set frequencies and serve only as oracle references; ReCAP selects without them.
We also rescore ReCAP's histories with its learned heads in place
of the frequencies, which measures how well the heads estimate $f$ itself (Learned/Emp). Details are in Sec.~\ref{appendix:empirical_guarantees}.

\begin{table}[t]
\centering
\footnotesize
\setlength{\tabcolsep}{4pt}
\caption{{Fixed-budget objective on OAI.} Policy columns compare each policy's objective with the optimum under the same empirical $f$; Theorem~\ref{thm:budget} guarantees Exact greedy/OPT $\geq 1-1/e\approx0.632$. The head-accuracy column rescores ReCAP's own histories with its learned heads, measuring estimation error in $f$ rather than policy quality. For $k\geq6$ all columns are unchanged to three decimals. Full results in Table~\ref{tab:greedy_optimal_full}.}
\label{tab:greedy_optimal}
\begin{tabular}{cc|cc|c}
\toprule
& & \multicolumn{2}{c|}{\textbf{Policy / OPT}} & \textbf{Head accuracy} \\
$k$ & $f_{\mathrm{OPT}}$ & Exact greedy & ReCAP & Learned/Emp \\
\midrule
1 & 0.503 & 1.000 & 1.000 & 0.996 \\
2 & 0.591 & 0.999 & 0.998 & 0.975 \\
3 & 0.618 & 0.998 & 0.989 & 0.970 \\
4 & 0.623 & 0.999 & 0.994 & 0.965 \\
5 & 0.626 & 1.000 & 0.999 & 0.957 \\
6 & 0.626 & 1.000 & 1.000 & 0.956 \\
\bottomrule
\end{tabular}
\end{table}

Greedy stays within 0.25\% of OPT at every budget
(Table~\ref{tab:greedy_optimal}), far above the worst case permitted by
Theorem~\ref{thm:budget}. ReCAP, which selects using only its learned heads,
attains at least $0.989$ of OPT, with the largest gap at $k=3$. All policies
coincide at $k=1$, where there is a single decision, and at $k=M$, where each
acquires the full vector, so intermediate budgets carry the comparison. The
heads, however, do not reproduce the value of $f$: rescoring ReCAP's own
histories with them underestimates $f$ by a margin that grows with $k$ and
settles at 4.4\% (Learned/Emp). ReCAP's near-optimality therefore comes from
ranking candidates correctly rather than from calibrated estimates of $f$. This is all its selection rule requires, and it suggests that the uniform absolute-error assumption of Proposition~\ref{prop:approx} is conservative in practice.




\section{Conclusion}
\label{sec:conclusion}
We have shown that in sequential concept acquisition, a correct diagnosis does not imply a reliable explanation, and that this gap can be closed by design: ReCAP’s label-discriminating objective, which accounts for concept-predictor errors and abstentions, yields shorter explanations that are markedly more reliable without sacrificing accuracy. More broadly, our results suggest that concept-based models should be evaluated on whether their explanations are correct, not only on whether their diagnoses are, and RCA offers a simple way to do so. ReCAP’s advantage is largest when findings are complementary, and its guarantees concern fixed budgets; extending them to stopping at a target error, and validating explanation reliability with clinicians, are natural next steps.

\bibliographystyle{unsrt}  
\bibliography{references}  

\appendix
\clearpage
\section{Proofs - Diminishing Returns and Fidelity}
\label{appendix:proofs}
Proposition~\ref{prop:adasub} is the adaptive monotonicity and submodularity of equivalence-class edge cutting~\cite{golovin2010near}, with CP-reported states as hypotheses and diagnoses as classes. We give a self-contained proof that works directly with real-valued probability masses, avoiding the reduction to rational priors, and bounds the partial derivatives of the gain directly.

\paragraph{Proposition~\ref{prop:adasub} (Diminishing returns).}
\begin{proof}
Let $\mathcal{H}$ be reachable, so that $p(\mathcal{H}) > 0$ and the
conditional answer probabilities below are defined. By
Assumption~\ref{ass:determinism} every
state $v \in V(\mathcal{H})$ has a single value $v_m \in \mathcal{A}$, so
querying $m$ \emph{partitions} the version space,
\begin{equation}
\label{eq:partition}
V(\mathcal{H}) \;=\; \biguplus_{a \in \mathcal{A}} V\big(\mathcal{H} \cup \{(m,a)\}\big).
\end{equation}
Write, for $a \in \mathcal{A}$ and $y \in \mathcal{Y}$,
\begin{align*}
s_a(y) &\;\triangleq\; p\big(\mathcal{H} \cup \{(m,a)\},\, y\big), \\
S_a &\;\triangleq\; \textstyle\sum_y s_a(y), \qquad
S \;\triangleq\; \textstyle\sum_a S_a = p(\mathcal{H}), \\
P(y) &\;\triangleq\; \textstyle\sum_a s_a(y) = p(\mathcal{H}, y),
\end{align*}
where the last two identities follow from
Equation~\eqref{eq:partition}. Put
\begin{align*}
E_{ab} &\;\triangleq\; \sum_{y \neq y'} s_a(y)\, s_b(y'), \\
A_a &\;\triangleq\; E_{aa} = R\big(\mathcal{H} \cup \{(m,a)\}\big),
\end{align*}
so that $R(\mathcal{H}) = \sum_{y \neq y'} P(y)P(y') = \sum_{a,b} E_{ab}$ and
$p(a \mid \mathcal{H}) = S_a/S$. A pair counted in $E_{ab}$ with $a \neq b$ has
its endpoints in different branches and is cut by querying $m$ whatever the
answer, whereas $A_a$ collects the pairs that survive when the answer is $a$.
 
Since $f = R(\varnothing) - R$, the constant cancels, and separating the
diagonal terms of the double sum from the rest gives
\begin{align}
\Delta(m \mid \mathcal{H})
&= R(\mathcal{H}) - \sum_a p(a \mid \mathcal{H})\, A_a \nonumber\\
&= \sum_{a,b} E_{ab} - \sum_a \tfrac{S_a}{S}\,A_a \nonumber\\
&= \sum_{a \neq b} E_{ab} + \sum_a A_a\Big(1 - \tfrac{S_a}{S}\Big) \nonumber \\
&= \sum_{a \neq b} E_{ab} + \frac{1}{S}\sum_{a \neq b} A_a S_b \nonumber \\
&= \sum_{a \neq b} \Big[ E_{ab} + \frac{A_a S_b}{S} \Big].
\label{eq:gain_exact}
\end{align}
 
\vspace{2pt}\noindent\textbf{Adaptive monotonicity.}
Every term in Equation~\eqref{eq:gain_exact} is a product of non-negative
quantities, so $\Delta(m \mid \mathcal{H}) \geq 0$ for every such
$\mathcal{H}$ and $m$.
 
\vspace{2pt}\noindent\textbf{Coordinatewise monotonicity.}
Collect the branch--label masses into the non-negative vector
$\mathbf{s} = \big(s_a(y)\big)$, indexed by $a \in \mathcal{A}$ and
$y \in \mathcal{Y}$, and regard the gain as a function
$\Delta(\mathbf{s}) = R - T/S$, where $T \triangleq \sum_a S_a A_a$. Fix
$u \in \mathcal{A}$ and $z \in \mathcal{Y}$, and write
$q_u \triangleq S_u - s_u(z)$ and $Q \triangleq S - P(z)$ for the masses
carrying a label other than $z$, within branch $u$ and within
$V(\mathcal{H})$ respectively. Differentiating,
\begin{equation*}
\frac{\partial R}{\partial s_u(z)} = 2Q, \qquad
\frac{\partial S}{\partial s_u(z)} = 1,
\end{equation*}
\begin{equation*}
\frac{\partial T}{\partial s_u(z)} = A_u + 2 S_u q_u,
\end{equation*}
so that
\begin{equation}
\label{eq:partial}
S^2\, \frac{\partial \Delta}{\partial s_u(z)}
\;=\; 2QS^2 - S\big(A_u + 2S_u q_u\big) + T .
\end{equation}
We bound the right-hand side using three elementary facts. First,
$T \geq S_u A_u$, by dropping the other non-negative terms. Second,
\begin{equation*}
A_u \;=\; S_u^2 - \sum_y s_u(y)^2 \;\leq\; S_u^2 - s_u(z)^2 \;\leq\; 2 S_u q_u,
\end{equation*}
the last step being $\big(S_u - s_u(z)\big)^2 \geq 0$. Third, $Q \geq q_u$,
since the mass of labels other than $z$ in $V(\mathcal{H})$ is at least that
within branch $u$. Applying these in turn to Equation~\eqref{eq:partial},
\begin{align*}
S^2 \frac{\partial \Delta}{\partial s_u(z)}
&\;\geq\; 2QS^2 - (S - S_u) A_u - 2 S S_u q_u \nonumber \\
&\;\geq\; 2QS^2 - 2 S_u q_u\big[(S - S_u) + S\big] \\
&\;=\; 2QS^2 - 2 S_u q_u (2S - S_u) \\
&\;\geq\; 2 q_u \big( S^2 - 2 S S_u + S_u^2 \big) \\
&\;=\; 2 q_u (S - S_u)^2 \;\geq\; 0 .
\end{align*}
Hence, if any branch-label mass $s_a(y)$ is increased while the others are held
fixed, the marginal gain $\Delta$ cannot decrease.
 
\vspace{2pt}\noindent\textbf{Adaptive submodularity.}
Let $\mathcal{H}'$ be a consistent extension of $\mathcal{H}$ with
$m \notin \mathrm{dom}(\mathcal{H}')$, and let $\mathbf{s}, \mathbf{s}'$ be the
branch-mass vectors of $\mathcal{H}, \mathcal{H}'$. Since
$V(\mathcal{H}') \subseteq V(\mathcal{H})$, for every $a$ we have
$V(\mathcal{H}' \cup \{(m,a)\}) \subseteq V(\mathcal{H} \cup \{(m,a)\})$,
and therefore $s'_a(y) \leq s_a(y)$ for all $a$ and $y$. As proved
above, $\Delta$ is non-decreasing in each coordinate on the non-negative orthant
(where $S>0$), so moving from $\mathbf{s}'$ to $\mathbf{s}$ one coordinate at a time gives
\begin{equation*}
\Delta(m \mid \mathcal{H}') = \Delta(\mathbf{s}')
\;\leq\; \Delta(\mathbf{s}) = \Delta(m \mid \mathcal{H}),
\end{equation*}
which is adaptive submodularity.
\end{proof}

\paragraph{Proposition~\ref{prop:sandwich}.}
This is the pointwise form of the Cover--Hart bound~\cite{cover1967nearest}, from which their bound on the asymptotic nearest-neighbour risk follows by averaging over $x$.
\begin{proof}
Let $\beta=\max_y b_y$, attained at $y^\ast$, so $B(b)=1-\beta$. Since $b_y\leq\beta$ for every $y$, $\sum_y b_y^2\leq\beta\sum_y b_y=\beta$, hence $G(b)\geq 1-\beta=B(b)$. For the upper bound, the remaining mass $\sum_{y\neq y^\ast}b_y=1-\beta$ is spread over $|\mathcal{Y}|-1$ labels, so by the Cauchy--Schwarz inequality $\sum_{y\neq y^\ast}b_y^2\geq(1-\beta)^2/(|\mathcal{Y}|-1)$. Therefore
\begin{align*}
G(b) &\leq 1-\beta^2-\tfrac{(1-\beta)^2}{|\mathcal{Y}|-1}\\
&= B(2-B)-\tfrac{B^2}{|\mathcal{Y}|-1}\\
&= B\Big(2-\tfrac{|\mathcal{Y}|}{|\mathcal{Y}|-1}B\Big),
\end{align*}
writing $B=B(b)$.
\end{proof}

\paragraph{Proposition~\ref{prop:decomp} (Fidelity decomposition).}
\begin{proof}
We first show that the population Bayes posterior at the terminal history
equals the conditional expectation of the full-image posterior. Write
$P(y\mid\mathcal{H})\triangleq\Pr(Y=y\mid\mathcal{H})$. Under
Assumption~\ref{ass:determinism} and determinism of $\pi$, the terminal history
is a function of $x$, so $\sigma(\mathcal{H})\subseteq\sigma(x)$ and, by the
tower property, $\mathbb{E}[P(y\mid x)\mid\mathcal{H}]=P(y\mid\mathcal{H})$.
Let $h$ be any terminal history with positive probability. If $\mathcal{H}=h$
then $\hat c\in V(h)$. Conversely, if $\hat c\in V(h)$, then $\pi$ makes the
same decision as on $h$ at every prefix of $h$, since its query and stopping
decisions depend only on acquired outcomes, so it again terminates at $h$.
Hence $\{\mathcal{H}=h\}=\{\hat c\in V(h)\}$ and
\begin{align}
P(y\mid\mathcal{H}=h) &=\Pr\big(Y=y\mid\hat c\in V(h)\big) \\&=\frac{p(h,y)}{p(h)}=p(y\mid h).
\end{align}
Summing over $y \in \mathcal{Y}$ and using
$\sum_y b_y^2 = 1 - G_x$ together with
$\sum_y \bar b_y^2 = 1 - G(\mathcal{H})$,
\begin{align*}
\mathbb{E}\big[ d_{\mathrm{B}} \big]
&= \mathbb{E}\Big[\textstyle\sum_y b_y^2\Big] - \mathbb{E}\Big[\textstyle\sum_y \bar b_y^2\Big] \\
&= \mathbb{E}[1 - G_x] - \mathbb{E}\big[1 - G(\mathcal{H})\big] \\
&= \mathbb{E}\big[G(\mathcal{H})\big] - \mathbb{E}[G_x],
\end{align*}
which is Equation~\eqref{eq:fidelity_identity}.
\end{proof}
 
\paragraph{Corollary~\ref{cor:sufficiency}.}
\begin{proof}
By the first part of the proof of Proposition~\ref{prop:decomp},
$P(y\mid\mathcal{H})=p(y\mid\mathcal{H})$, so the conditional misclassification
probability of the MAP diagnosis at a terminal history is
\begin{align*}
P\big(\hat y(\mathcal{H}) \neq y \mid \mathcal{H}\big)
&= 1 - \max_y p(y \mid \mathcal{H})\\
&= B\big(p(\cdot \mid \mathcal{H})\big)
\leq G(\mathcal{H}),
\end{align*}
the inequality by Proposition~\ref{prop:sandwich}. Since ties are broken by a
fixed rule, $\hat y(\mathcal{H})$ is a function of $\mathcal{H}$. Taking
expectations over terminal histories and using the tower property and the
hypothesis $\mathbb{E}[G(\mathcal{H})]\leq\delta'$,
\begin{equation}
\label{eq:risk_bound}
P\big(\hat y(\mathcal{H}) \neq y\big)
= \mathbb{E}\big[ P\big(\hat y(\mathcal{H}) \neq y \mid \mathcal{H}\big) \big]
\leq \mathbb{E}[G(\mathcal{H})] \leq \delta'.
\end{equation}
For fidelity, Equation~\eqref{eq:fidelity_identity} and $G_x\geq0$ give
\begin{align*}
\mathbb{E}_x\big[ d_{\mathrm{B}}\big(P(y \mid x), p(y \mid \mathcal{H})\big) \big]
&= \mathbb{E}[G(\mathcal{H})] - \mathbb{E}[G_x]\\
&\leq \mathbb{E}[G(\mathcal{H})] \leq \delta',
\end{align*}
so the fidelity constraint of Equation~\eqref{eq:base} holds with
$d=d_{\mathrm{B}}$ and $\delta=\delta'$.
\end{proof}
The hypothesis holds in particular when every terminal history satisfies
$G(\mathcal{H})\leq\delta'$, but it also covers policies that sometimes stop
before reaching this threshold, for example at the query cap of
Algorithm~\ref{alg:greedy}, provided the average terminal Gibbs error stays
within $\delta'$. Such policies satisfy the conditions of
Proposition~\ref{prop:decomp}, since the cap depends only on $|\mathcal{H}|$.

\section{Supplementary Guarantees} \label{appendix:guarantees} 
This section provides the proofs and qualifications underlying the acquisition, reliability, and fidelity-floor guarantees. The acquisition results compare expected objective value under a fixed query budget, rather than directly bounding the number of queries required to attain a fidelity threshold. They concern acquisition histories before pruning. \subsection{Population Greedy Guarantee} The population rule evaluates marginal gains under $p(\hat c,y)$ and selects an exact maximiser. Its comparison class comprises adaptive policies observing the same CP outputs, with at most $k$ queries almost surely. Neither the greedy policy nor its comparator has access to $c^\star$. \paragraph{Theorem~\ref{thm:budget} (Budgeted guarantee).} \begin{proof} We apply the adaptive greedy theorem of~\cite{golovin2011adaptive}. The ground set is $\{1,\dots,M\}$, the realisation is $\hat c\in\mathcal{A}^M$, and querying item $m$ reveals $\hat c_m$. Under Assumption~\ref{ass:determinism}, observations are deterministic given the realisation. The objective $f(\mathcal{H})=R(\varnothing)-R(\mathcal{H})$ is bounded and normalised, with $f(\varnothing)=0$. It depends only on observed coordinates and is adaptively monotone and adaptively submodular by Proposition~\ref{prop:adasub}. Equation~\eqref{eq:greedy_rule} selects an exact maximiser of its conditional expected marginal gain. The adaptive greedy theorem therefore yields Equation~\eqref{eq:budget_bound}. \end{proof} The guarantee applies to the population greedy rule run for $k$ acquisitions, without fidelity-based early stopping. It is a worst-case comparison of objective values, not a guarantee that greedy acquisition attains a specified fidelity threshold within $k$ queries. \subsection{Learned Scores and Approximate Gains} The deployed rule minimises the learned score $s_\eta(m\mid\mathcal{H})$ (Sec.~\ref{subsec:algorithm}), which is equivalent to maximising the absolute estimate \[ \widehat\Delta(m\mid\mathcal{H}) =p(\mathcal{H})^2\big(G_\eta(\mathcal{H})-s_\eta(m\mid\mathcal{H})\big), \] since $p(\mathcal{H})^2$ and $G_\eta(\mathcal{H})$ are common to all candidates. The deployed rule therefore maximises $\widehat\Delta$ without computing $p(\mathcal{H})$. Enumerating all candidates eliminates optimisation error over the learned scores, but not estimation error relative to the population gains. The following proposition quantifies the latter. \begin{proposition}[Guarantee under approximate gains] \label{prop:approx} Suppose \[ \big| \widehat\Delta(m\mid\mathcal{H}) -\Delta(m\mid\mathcal{H}) \big|\leq\epsilon \] for every positive-probability history reachable within $k$ steps and every unacquired concept $m$. Then the policy $\hat\pi$ selecting $\arg\max_m\widehat\Delta(m\mid\mathcal{H})$ for $k$ acquisitions satisfies \begin{equation} \label{eq:approx_bound} f_{\mathrm{avg}}(\hat\pi^{[k]}) \geq \left(1-\frac{1}{e}\right) \max_{\pi:\,|\tau^\pi|\leq k}f_{\mathrm{avg}}(\pi) -2k\epsilon. \end{equation} \end{proposition} \begin{proof} Let \[ \mathrm{OPT} =\max_{\pi:\,|\tau^\pi|\leq k}f_{\mathrm{avg}}(\pi), \qquad f_t=f_{\mathrm{avg}}(\hat\pi^{[t]}). \] At history $\mathcal{H}_t$, let $\hat m$ maximise $\widehat\Delta$ and $m^\star$ maximise $\Delta$. The uniform error bound gives \begin{align*} \Delta(\hat m\mid\mathcal{H}_t) &\geq\widehat\Delta(\hat m\mid\mathcal{H}_t)-\epsilon\\ &\geq\widehat\Delta(m^\star\mid\mathcal{H}_t)-\epsilon\\ &\geq\Delta(m^\star\mid\mathcal{H}_t)-2\epsilon. \end{align*} By adaptive monotonicity and adaptive submodularity, the adaptive greedy comparison of~\cite{golovin2011adaptive} yields \[ \mathrm{OPT}-f_t \leq k\,\mathbb{E}\Big[ \max_m\Delta(m\mid\mathcal{H}_t) \Big]. \] Consequently, \[ f_{t+1}-f_t \geq\frac{\mathrm{OPT}-f_t}{k}-2\epsilon. \] Writing $d_t=\mathrm{OPT}-f_t$, we obtain \[ d_{t+1}\leq \left(1-\frac{1}{k}\right)d_t+2\epsilon. \] Since $f_0=0$, unrolling the recurrence gives \begin{align*} d_k &\leq \left(1-\frac{1}{k}\right)^k\mathrm{OPT} +2\epsilon \sum_{t=0}^{k-1} \left(1-\frac{1}{k}\right)^t\\ &\leq \frac{1}{e}\mathrm{OPT}+2k\epsilon. \end{align*} Rearranging proves Equation~\eqref{eq:approx_bound}. \end{proof} The assumption controls absolute marginal-gain error uniformly over relevant histories and candidates. It is not implied by exhaustive candidate evaluation or by low average training loss. Accordingly, we do not assume $\epsilon=0$. \paragraph{Early stopping and pruning.} Proposition~\ref{prop:approx} concerns fixed-horizon acquisition. It does not automatically provide the same approximation factor for trajectories stopped early by the learned fidelity criterion. Corollary~\ref{cor:sufficiency} supplies a separate guarantee on diagnostic risk and fidelity whenever the expected terminal Gibbs error under the population Bayes posterior is at most $\delta'$, regardless of how acquisition stopped. Applying that guarantee to $p_\eta$ requires additional control of posterior estimation error. Post-acquisition pruning leaves the query trajectory unchanged, so the acquisition bounds remain statements about the unpruned history. They need not hold for the surrogate value evaluated on $\mathcal{H}^\dagger$. \subsection{Reliability under Adaptive Selection} Let $N^\pi_{\mathrm{a}}$ denote the number of signed (non-$\bot$) assertions in the trajectory of a policy $\pi$, and $\rho^\pi$ the number of incorrect ones. Because the policy selects concepts adaptively, selective reliability must hold conditionally on the acquired history.
\begin{assumption}[Selective reliability]
\label{ass:selective}
There exists $\varepsilon_\kappa\in[0,1]$ such that, for every positive-probability history $\mathcal{H}$ and every $m\notin\mathrm{dom}(\mathcal{H})$ with $\Pr(\hat c_m\neq\bot\mid\hat c\in V(\mathcal{H}))>0$,
$\Pr\big(\hat c_m\neq c^\star_m\mid\hat c\in V(\mathcal{H}),\,\hat c_m\neq\bot\big)\leq\varepsilon_\kappa$.
\end{assumption}
By Equation~\eqref{eq:admissible}, $\hat c_m\neq\bot$ is equivalent to $\sigma_m\leq\kappa$.
\begin{proposition}[Uniform reliability]
\label{prop:reliability}
Under Assumptions~\ref{ass:determinism} and~\ref{ass:selective}, every policy $\pi$ as in Proposition~\ref{prop:decomp} satisfies $\mathbb{E}[\rho^\pi]\leq\varepsilon_\kappa\,\mathbb{E}[N^\pi_{\mathrm{a}}]\leq\varepsilon_\kappa\,\mathbb{E}\big[|\tau^\pi|\big]$. In particular, Algorithm~\ref{alg:greedy} satisfies $\mathbb{E}[\rho^\pi]\leq\varepsilon_\kappa\gamma_{\max}$.
\end{proposition}
The bound controls incorrect assertions independently of the quality of the selection rule.
\begin{proof}
Let $T=|\tau^\pi(x)|$ and $\mathcal{F}_{k-1}=\sigma(\mathcal{H}_{1:k-1})$, with stopped histories held fixed after termination, and define $j_k$ arbitrarily after termination. Since $\pi$ is deterministic and its query and stopping decisions depend only on acquired outcomes, the event $\{k\leq T\}$ and the selected concept $j_k$ are $\mathcal{F}_{k-1}$-measurable. Moreover, for any history $h$ of length $k-1$ at which $\pi$ has not stopped, every $\hat c\in V(h)$ reproduces $h$ and does not stop, as in the proof of Proposition~\ref{prop:decomp}, so
\begin{equation*}
\{\mathcal{H}_{1:k-1}=h,\;k\leq T\}=\{\hat c\in V(h)\}.
\end{equation*}
The assertion event $\{\hat c_{j_k}\neq\bot\}$ is not generally $\mathcal{F}_{k-1}$-measurable, since it is revealed only upon querying. Set
\[
A_k=\mathds{1}[k\leq T]\,\mathds{1}[\hat c_{j_k}\neq\bot],
\qquad
E_k=\mathds{1}[\hat c_{j_k}\neq c^\star_{j_k}].
\]
On $\{k\leq T\}$, conditioning on $\mathcal{F}_{k-1}$ is conditioning on $\hat c\in V(h)$ with $j_k=j_k(h)\notin\mathrm{dom}(h)$ fixed, so Assumption~\ref{ass:selective} gives
\begin{align*}
\mathbb{E}[A_kE_k\mid\mathcal{F}_{k-1}]
&=\mathds{1}[k\leq T]\,\Pr(\hat c_{j_k}\neq\bot\mid\mathcal{F}_{k-1})\\
&\quad\times\Pr(E_k=1\mid\mathcal{F}_{k-1},\hat c_{j_k}\neq\bot)\\
&\leq\varepsilon_\kappa\,\mathbb{E}[A_k\mid\mathcal{F}_{k-1}].
\end{align*}
If the assertion event has zero conditional probability, both sides are zero. Taking expectations and summing over query steps yields
\[
\mathbb{E}[\rho^\pi]=\sum_k\mathbb{E}[A_kE_k]\leq\varepsilon_\kappa\sum_k\mathbb{E}[A_k]=\varepsilon_\kappa\mathbb{E}[N^\pi_{\mathrm{a}}].
\]
Since $N^\pi_{\mathrm{a}}\leq T$, $\mathbb{E}[\rho^\pi]\leq\varepsilon_\kappa\mathbb{E}[T]$. For Algorithm~\ref{alg:greedy}, $T\leq\gamma_{\max}$ by the query cap, so $\mathbb{E}[\rho^\pi]\leq\varepsilon_\kappa\gamma_{\max}$.
\end{proof} 
The bound applies to Algorithm~\ref{alg:greedy} under fixed tie-breaking and does not depend on its approximation ratio. Pruning cannot increase the number of incorrect assertions, so the trajectory-level upper bound also covers the delivered explanation. This result is conditional on Assumption~\ref{ass:selective}. A marginal selective-risk estimate alone does not establish the required history-conditional bound. Moreover, reliability must be reported alongside assertion and refusal counts, since universal abstention produces no incorrect assertions but no signed concept explanation. \subsection{Characterisation of the Fidelity Floor} The fidelity floor measures information loss after all CP outputs have been observed. It is distinct from intrinsic label ambiguity, which remains even when the true concepts are available.
\begin{proposition}[Characterisation of the floor]
\label{prop:floor}
With $d=d_{\mathrm{B}}$ in Equation~\eqref{eq:delta_min} and under Assumptions~\ref{ass:sufficiency} and~\ref{ass:determinism},
\begin{equation}
\label{eq:floor_decomp}
\delta_{\min}=\mathbb{E}\big[G(P(\cdot\mid\hat c))\big]-\mathbb{E}\big[G(P(\cdot\mid c^\star))\big].
\end{equation}
\end{proposition}
\begin{proof}
Let $Z=P(\cdot\mid x)$ and $\bar Z=\mathbb{E}[Z\mid\hat c]$. Since $\hat c$ is a function of $x$ (Assumption~\ref{ass:determinism}), the tower property gives $\bar Z=P(\cdot\mid\hat c)$. For any predictor $q$ measurable with respect to $\hat c$, the tower and pull-out properties make the cross term vanish, so
\[
\mathbb{E}\|Z-q\|_2^2=\mathbb{E}\|Z-\bar Z\|_2^2+\mathbb{E}\|\bar Z-q\|_2^2 .
\]
Thus the infimum in Equation~\eqref{eq:delta_min} is attained at $q=\bar Z$, and by the same argument as in Proposition~\ref{prop:decomp},
\[
\delta_{\min}=\mathbb{E}\|Z\|_2^2-\mathbb{E}\|\bar Z\|_2^2=\mathbb{E}\big[G(P(\cdot\mid\hat c))\big]-\mathbb{E}[G_x].
\]
Under Assumption~\ref{ass:sufficiency}, $P(\cdot\mid x)=P(\cdot\mid c^\star)$ almost surely, so this is Equation~\eqref{eq:floor_decomp}. Equivalently,
\[
\delta_{\min}=\mathbb{E}\Big[\big\|P(\cdot\mid c^\star)-P(\cdot\mid\hat c)\big\|_2^2\Big]\geq0 .
\]
Equality holds precisely when $P(\cdot\mid c^\star)$ is $\hat c$-measurable, or equivalently when $\hat c$ preserves the label posterior available from the true concepts.
\end{proof} 
The intrinsic term $\mathbb{E}[G(P(\cdot\mid c^\star))]$ is independent of the CP and abstention threshold. Changes in $\delta_{\min}$ therefore enter through the ambiguity remaining given $\hat c$. Estimating either term requires estimating its conditional label distribution; concept annotations alone do not make these population quantities directly observable.

\section{Training and Explanation Pruning} 
\subsection{Training the Predictive Heads} \label{appendix:training} Training proceeds in three stages. First, we initialise the diagnostic head $p_\eta$ on complete histories using cross-entropy. Second, we jointly train both heads on partial histories generated by V-IP-style random history sampling \cite{chattopadhyayvariational}, drawing a fresh concept subset without replacement for each training sample. For a mini-batch $\mathcal{B}$, the losses are \begin{align} \mathcal{L}_{\mathrm{cls}} &=\frac{1}{|\mathcal{B}|} \sum_{i\in\mathcal{B}} \mathrm{CE}\big(p_\eta(\cdot\mid\mathcal{H}_i),y_i\big),\\ \mathcal{L}_{\mathrm{ans}} &= \frac{ \sum_{i\in\mathcal{B}}\sum_{m\in\mathcal{C}_i} \mathrm{CE}\big( p_\psi(\cdot\mid m,\mathcal{H}_i), c_{i,m} \big) }{ \sum_{i\in\mathcal{B}}|\mathcal{C}_i| }, \end{align} where $\mathcal{C}_i$ contains the unacquired concepts. Both heads minimise $\mathcal{L}_{\mathrm{cls}}+\mathcal{L}_{\mathrm{ans}}$. Finally, we fine-tune both heads on trajectories generated by Equation~\eqref{eq:greedy_rule}. For each mini-batch, complete trajectories are generated with fixed parameters. The same losses are then evaluated over all visited histories before a single parameter update. The predictive heads directly supply the acquisition scores; no separate querier or policy network is trained. 
\subsection{Pruning the Delivered Explanation} \label{appendix:pruning} A concept may separate states without changing the label posterior, particularly when its information is useful only in combination with another concept. If the complementary concept is refused or never acquired, the earlier query may contribute nothing to the terminal explanation. Because its eventual contribution is unknown during acquisition, we leave the selection rule unchanged and prune only the terminal history. An acquired concept $m$ is exactly \emph{inert} if \begin{equation} \label{eq:inert} p_\eta(\cdot\mid\mathcal{H}\setminus\{m\}) =p_\eta(\cdot\mid\mathcal{H}). \end{equation} Removing an inert concept preserves the full posterior, the MAP diagnosis, and the Gibbs error. Equality of Gibbs errors alone is insufficient: distinct posteriors can have equal Gibbs error but different MAP labels. The implemented criterion, Equation~\eqref{eq:inert_approx}, permits a posterior change of at most $\xi$ while explicitly preserving the diagnosis. The diagnosis check is necessary because even a small posterior shift can reverse closely ranked classes. MAP ties are resolved by a fixed rule. Removals are sequential, with each candidate tested against the current retained history. Simultaneous removal is unsafe: two duplicated concepts may each be individually inert, although removing both changes the posterior. Algorithm~\ref{alg:greedy} therefore removes one eligible concept at a time using a fixed order. Let $\mathcal{H}^\dagger$ denote the final retained history. Under exact pruning, its posterior equals that of the original terminal history. Under approximate pruning, the diagnosis is preserved at every removal. If $r$ concepts are removed, the triangle inequality gives \[ \big\| p_\eta(\cdot\mid\mathcal{H}^\dagger) -p_\eta(\cdot\mid\mathcal{H}) \big\|_1 \leq r\xi. \] Approximate pruning therefore preserves the diagnosis, but not necessarily the Gibbs error or the original stopping tolerance. Pruning changes neither the acquisition trajectory nor the comparison class of Sec.~\ref{subsec:guarantees}. A removed concept was still queried and may have influenced later selections. Accordingly, $|\tau^\pi(x)|-|\mathcal{H}^\dagger|$ counts acquired concepts omitted from the delivered explanation, rather than queries avoided. If the fidelity tolerance is unattainable, pruning removes only concepts satisfying Equation~\eqref{eq:inert_approx}. It need not remove every query made without attaining the threshold, since such queries may still change the posterior.

\section{Dataset Details}
\label{appendix:datasets}

We use five medical imaging benchmarks covering retinal fundus photography,
breast ultrasound, dermoscopy, knee radiography, and blood-smear microscopy.
When an official partition is available, we follow the provided split;
otherwise, samples are randomly divided into training, validation, and test
sets with a 70/10/20 ratio. Table~\ref{tab:dataset_details} provides the
prediction targets and concept vocabulary used for each benchmark. Datasets
are ordered by increasing concept-space size.

\begin{table*}[!t]
\caption{\textbf{Dataset characteristics and concept vocabulary.}
The five benchmarks are arranged by increasing number of concepts:
DDR (4), BrEaST (7), PH2 (8), OAI (10), and WBCatt (29).
Concept abbreviations are shown in parentheses and are used
in the qualitative acquisition visualizations.}
\label{tab:dataset_details}
\centering
\footnotesize
\renewcommand{\arraystretch}{1.15}
\setlength{\tabcolsep}{4pt}
\begin{tabular}{p{0.08\textwidth}
                p{0.12\textwidth}
                p{0.17\textwidth}
                p{0.55\textwidth}}
\toprule
\textbf{Dataset} &
\textbf{Modality} &
\textbf{Prediction Classes} &
\textbf{Concept Vocabulary (Abbreviation)} \\
\midrule

DDR &
Fundus &
No DR, DR &
Hard exudates (EX), hemorrhages (HE),
microaneurysms (MA), soft exudates (SE)
\\

\midrule

BrEaST &
Ultrasound &
Benign, malignant &
Posterior features (PF), halo (HL),
calcifications (CAL), skin thickening (ST),
irregular shape (IS), not circumscribed margin (NCM),
heterogeneous or hyperechoic (HH)
\\

\midrule

PH2 &
Dermoscopy &
Nevus, melanoma &
Typical pigment network (TPN),
atypical pigment network (APN),
irregular streaks (IST),
regular streaks (RST),
regular dots and globules (RDG),
irregular dots and globules (IDG),
blue-whitish veil (BWV),
regression structures (RS)
\\

\midrule

OAI &
Knee X-ray &
KLG 0, KLG 1, KLG 2, KLG 3 &
Osteophytes (Femur, Medial; OFM),
sclerosis (Femur, Medial; SFM),
joint-space narrowing (Medial; JSM),
osteophytes (Tibia, Medial; OTM),
sclerosis (Tibia, Medial; STM),
osteophytes (Femur, Lateral; OFL),
sclerosis (Femur, Lateral; SFL),
joint-space narrowing (Lateral; JSL),
osteophytes (Tibia, Lateral; OTL),
sclerosis (Tibia, Lateral; STL)
\\

\midrule

WBCatt &
Microscopy &
Neutrophil, eosinophil, basophil, monocyte, lymphocyte &
Granularity (GRAN),
cytoplasm vacuole (CVAC),
cell size big (CS-B),
cell size small (CS-S),
cell shape irregular (SH-I),
cell shape round (SH-R),
nucleus shape irregular (NS-I),
nucleus shape segmented-bilobed (NS-B),
nucleus shape segmented-multilobed (NS-M),
nucleus shape unsegmented-band (NS-UB),
nucleus shape unsegmented-indented (NS-UI),
nucleus shape unsegmented-round (NS-UR),
nuclear--cytoplasmic ratio high (NCR-H),
nuclear--cytoplasmic ratio low (NCR-L),
chromatin density densely (CH-D),
chromatin density loosely (CH-L),
cytoplasm texture clear (CT-C),
cytoplasm texture frosted (CT-F),
cytoplasm colour blue (CC-B),
cytoplasm colour light blue (CC-LB),
cytoplasm colour purple blue (CC-PB),
granule type coarse (GT-C),
granule type nil (GT-N),
granule type round (GT-R),
granule type small (GT-S),
granule colour nil (GC-N),
granule colour pink (GC-P),
granule colour purple (GC-PU),
granule colour red (GC-R)
\\

\bottomrule
\end{tabular}
\end{table*}

\section{Implementation Details}
\label{appendix:implementation}

\paragraph{Network Architecture.}
The diagnosis predictor $p_{\eta}(y\mid H)$ and
concept-answer predictor $p_{\psi}(a\mid m,H)$
share a two-layer MLP encoder, followed by
separate task-specific prediction heads.
The shared encoder contains two fully connected
layers with 256 hidden units, each followed by
a ReLU activation and dropout with probability
$0.2$. Each prediction head contains a fully
connected layer with 128 hidden units, followed
by ReLU and dropout.
The diagnosis head outputs $C$ class logits,
whereas the answer head outputs $3M$ logits,
reshaped into an $M\times 3$ tensor representing
the three possible concept-answer outcomes:
negative ($-1$), positive ($+1$), and
abstention ($\bot$).

\paragraph{Variable-Length History Representation.}
To accommodate partial acquisition histories
of different lengths, we use a fixed-dimensional
representation with an explicit query mask.
For each history $H$, we construct a
concept-answer vector
$\mathbf{h}\in\{-1,0,+1,2\}^{M}$ and a binary
query mask $\mathbf{b}\in\{0,1\}^{M}$.
The entries of $\mathbf{h}$ indicate observed
positive ($+1$), negative ($-1$), abstained
($2$), and unqueried ($0$) concepts.
The query mask is defined as
$b_m=\mathbb{I}[h_m\neq 0]$.

The diagnosis and answer predictors use
different views of the same observed history.
For the answer predictor, abstentions are
explicitly retained:
\begin{equation}
\mathbf{z}_{\psi}(H)
=
[\mathbf{h};\mathbf{b}]
\in\mathbb{R}^{2M}.
\end{equation}
For the diagnosis predictor, abstained
concepts are treated as unqueried.
Specifically, we define
\begin{equation}
\widetilde{h}_m =
\begin{cases}
0, & h_m=2,\\
h_m, & \text{otherwise},
\end{cases}
\end{equation}
with
$\widetilde{b}_m=
\mathbb{I}[\widetilde{h}_m\neq 0]$.
The diagnosis input is
\begin{equation}
\mathbf{z}_{\eta}(H)
=
[\widetilde{\mathbf{h}};
\widetilde{\mathbf{b}}]
\in\mathbb{R}^{2M}.
\end{equation}
Thus, the answer predictor retains information
about previous refusals, whereas the diagnosis
predictor uses only accepted concept evidence.
The fixed-dimensional representation allows
histories of different lengths to be processed
within the same mini-batch without sequence
padding.

Previously queried concepts are excluded from
subsequent acquisition. At each step, the
candidate set contains all concepts that have
not yet been queried:
\begin{equation}
\mathcal{A}(H)
=
\left\{
m\in\{1,\ldots,M\}:b_m=0
\right\}.
\end{equation}
No concept is excluded beforehand based on
its uncertainty or admissibility status.
Instead, the concept predictor determines
whether the selected query returns an accepted
answer or abstains.

\paragraph{Partial-History Sampling.}
The construction of Stage~2 (described in Sec.~\ref{appendix:training}) histories
depends on the size of the concept space.
For DDR, BrEaST, and PH2, we use exhaustive
enumeration of partial concept histories.
For OAI and WBCatt, we adopt the V-IP-style
random sampling strategy
\cite{chattopadhyayvariational}
to generate partial histories of different
lengths.

For random sampling, the history length
is sampled from the supported range,
and concepts are selected uniformly
without replacement. Training histories
are regenerated during training, whereas
validation histories remain fixed for
consistent checkpoint selection.

Stage~2 histories are sampled from the
complete concept space without excluding
concepts according to their uncertainty.
The observed concept responses are then
constructed using the selected history
indices and the corresponding concept
predictor outputs, including abstentions.

The maximum Stage~2 history length is
determined independently of the sequential
acquisition budget. The fixed-dimensional
history representation allows histories
of different lengths to be used within
the same training procedure.

\paragraph{Optimization and Model Selection.}
All three training stages use AdamW with
weight decay $10^{-5}$. The learning rates
are $10^{-3}$, $5\times10^{-4}$, and
$10^{-4}$ for Stages~1, 2, and 3,
respectively.

The three stages use separate epoch budgets
and mini-batch sizes, summarized in
Table~\ref{tab:supp_training_config}.
The selected Stage~1 checkpoint initializes
Stage~2, and the selected Stage~2 checkpoint
initializes Stage~3. For the acquisition
strategy ablation, each Stage~3 policy
is initialized from the same Stage~2
checkpoint.

Hyperparameters and checkpoints are selected
using the validation set, with diagnosis
Macro-F1 used for checkpoint selection.
The test set is reserved for final evaluation.

\begin{table*}[!t]
\centering
\caption{
\textbf{Dataset-specific implementation configurations.}
$M$ denotes the number of concepts.
$E_s$ and $B_s$ denote the training epochs
and mini-batch size for Stage~$s$, respectively.
$Q_{\max}$ denotes the maximum sequential
acquisition budget, $\delta'$ is the Gibbs
stopping tolerance, and $\xi$ is the terminal
pruning threshold.
Concept-wise uncertainty thresholds determine
the observed answer outcomes but do not
restrict the candidate set before acquisition.
}
\label{tab:supp_training_config}

\scriptsize
\setlength{\tabcolsep}{4pt}
\renewcommand{\arraystretch}{1.12}

\begin{tabular}{@{}lccccccccccc@{}}
\toprule
\textbf{Dataset}
& $\boldsymbol{M}$
& $\boldsymbol{E_1}$
& $\boldsymbol{E_2}$
& $\boldsymbol{E_3}$
& $\boldsymbol{B_1}$
& $\boldsymbol{B_2}$
& $\boldsymbol{B_3}$
& $\boldsymbol{Q_{\max}}$
& $\boldsymbol{\delta'}$
& $\boldsymbol{\xi}$
& \textbf{History Sampling} \\
\midrule

DDR
& 4 & 100 & 100 & 100
& 256 & 1024 & 128
& 4 & 0.05 & 0.05 & Exhaustive \\

BrEaST
& 7 & 50 & 50 & 50
& 64 & 512 & 32
& 7 & 0.05 & 0.05 & Exhaustive \\

PH2
& 8 & 100 & 100 & 100
& 256 & 1024 & 128
& 8 & 0.05 & 0.05 & Exhaustive \\

OAI
& 10 & 100 & 100 & 40
& 1024 & 1024 & 1024
& 10 & 0.10 & 0.10 & Random \\

WBCatt
& 29 & 100 & 300 & 20
& 1024 & 1024 & 1024
& 10 & 0.03 & 0.05 & Random \\

\bottomrule
\end{tabular}
\end{table*}

\paragraph{Sequential Acquisition and Terminal Pruning.}
During Stage~3 and inference, the acquisition
policy selects concepts from the set of
previously unqueried concepts.
No image-specific admissibility filtering
is applied before concept selection.
Acquisition terminates when the Gibbs
stopping criterion is satisfied, the
maximum query budget is reached, or
all concepts have been queried.
The dataset-specific values of
$\delta'$ and $Q_{\max}$ are reported
in Table~\ref{tab:supp_training_config}.

At each step, the answer predictor estimates
the distribution of possible responses for
each unqueried concept. The acquisition
objective uses these predicted distributions
to evaluate candidate queries. After
selection, the actual concept predictor
response is revealed:
\begin{equation}
a_m(x)=
\begin{cases}
\bot, & \sigma_m(x)>\kappa_m,\\
+1, & \sigma_m(x)\leq\kappa_m
\ \text{and}\ \mu_m(x)\geq 0.5,\\
-1, & \text{otherwise}.
\end{cases}
\end{equation}
Here, $\mu_m(x)$ denotes the concept
predictor output, $\sigma_m(x)$ denotes
its uncertainty, and $\kappa_m$ is the
concept-specific uncertainty threshold.

The observed response is appended to
the acquisition history.


Terminal pruning is applied after acquisition without
retraining the prediction heads or changing the query
trajectory, as detailed in Sec.~\ref{appendix:pruning}.
RCA is computed before and after pruning using the
same eligible test trajectories, each containing at
least one originally accepted concept answer.

\paragraph{Greedy Acquisition and Pruning Algorithm}
Algorithm~\ref{alg:greedy} summarizes the complete greedy concept
acquisition and post-hoc pruning procedure described in the main paper.
\label{appendix:greedy_algorithm}
\begin{algorithm}[t] 
\caption{Greedy reliable concept acquisition} 
\label{alg:greedy} 
\begin{algorithmic}[1] 
\Require query access to $\hat c_m(x)$; heads $p_\psi,p_\eta$; tolerance $\delta'$; cap $\gamma_{\max}$; pruning tolerance $\xi$ 
\State $\mathcal{H}\gets\varnothing$; $\mathcal{C}\gets\{1,\dots,M\}$ 
\While{$G_\eta(\mathcal{H})>\delta'$ \textbf{and} $|\mathcal{H}|<\gamma_{\max}$ \textbf{and} $\mathcal{C}\neq\varnothing$} \For{each $m\in\mathcal{C}$} 
\State Compute $s_\eta(m\mid\mathcal{H})$ using Eq.~(\ref{equ:learned_score}).
\EndFor 
\State $m^\star\gets\arg\min_{m\in\mathcal{C}}s_\eta(m)$ \Comment{fixed tie-breaking} 
\State $a^\star\gets\hat c_{m^\star}(x)$ \Comment{may return $\bot$} 
\State $\mathcal{H}\gets \mathcal{H}\cup\{(m^\star,a^\star)\}$ ; $\mathcal{C}\gets\mathcal{C}\setminus\{m^\star\}$ \EndWhile 
\While{some $m\in\mathrm{dom}(\mathcal{H})$ satisfies Equation~\eqref{eq:inert_approx}} 
\State Remove one such $m$ from $\mathcal{H}$ using a fixed order 
\EndWhile 
\State 
\Return $\mathcal{H}$, $\hat y(\mathcal{H})$ 
\end{algorithmic} 
\end{algorithm} 

\section{Detailed Experimental Results}
\label{sec:supp_detailed_results}

Table~\ref{tab:main_full_results}
presents the detailed experimental
results across five medical imaging
datasets, complementing the cross-dataset
comparison in Table~\ref{tab:main_results}
of the main paper. We report Reliable
Concept Acquisition Rate (RCA),
Macro-F1 (MF1), accuracy (ACC),
and the average number of concepts
($Q_{\mathrm{avg}}$).
Results are expressed as mean $\pm$
standard deviation across three random
seeds. For ReCAP, results are reported both
before and after terminal pruning.

\begin{table*}[t]
\centering
\caption{
\textbf{Comparison across five medical imaging datasets.}
Results are reported as mean $\pm$ SD over three seeds. RCA measures concept reliability,
while MF1 and ACC measure diagnostic performance. Best and second-best results among deployable methods
are highlighted in \textbf{bold} and
\underline{underlined}, respectively.
GISS-U is a ground-truth-informed diagnostic reference
that selects candidate histories using the true
diagnosis while retaining potentially incorrect
concept predictions. It is highlighted in gray
and excluded from method ranking.
}
\label{tab:main_full_results}

\scriptsize
\setlength{\tabcolsep}{2.5pt}
\renewcommand{\arraystretch}{0.82}

\begin{tabular}{@{}lllcccc@{}}
\toprule
&
&
&
\textbf{Reliability}
&
\multicolumn{2}{c}{\textbf{Diagnosis Performance}}
&
\textbf{Efficiency} \\
\cmidrule(lr){4-4}
\cmidrule(lr){5-6}
\cmidrule(lr){7-7}

\textbf{Dataset}
& \textbf{Type}
& \textbf{Method}
& \textbf{RCA}$\uparrow$
& \textbf{MF1}$\uparrow$
& \textbf{ACC}$\uparrow$
& $\mathbf{Q}_{\mathrm{avg}}\downarrow$ \\
\midrule

\multirow{11}{*}{DDR}
& \multirow{3}{*}{Static}
& CBM~\cite{koh2020concept}
& 56.44$\pm$0.54
& 82.05$\pm$0.05
& 82.92$\pm$0.05
& 4.00$\pm$0.00 \\

& & ProbCBM~\cite{kim2023probabilistic}
& 57.67$\pm$0.34
& \textbf{83.21$\pm$0.02}
& 83.11$\pm$0.12
& 4.00$\pm$0.00 \\

& & Evi-CEM~\cite{gao2024evidential}
& 65.33$\pm$0.23
& 81.72$\pm$0.22
& 82.09$\pm$0.09
& 4.00$\pm$0.00 \\

\cmidrule(lr){2-7}

& \multirow{8}{*}{Sequential}
& Joint-AFA~\cite{shim2018joint}
& \underline{77.23$\pm$0.86}
& 82.31$\pm$0.08
& 82.92$\pm$0.11
& \textbf{1.49$\pm$1.85} \\

& & Generative IP~\cite{chattopadhyay2022interpretable}
& 61.39$\pm$0.00
& 82.29$\pm$0.00
& 83.17$\pm$0.00
& 3.69$\pm$0.00 \\

& & V-IP~\cite{chattopadhyayvariational}
& 56.44$\pm$1.05
& 82.32$\pm$0.05
& \underline{83.32$\pm$0.15}
& 4.00$\pm$0.00 \\

& & EUAV-IP~\cite{nahiduzzaman2026uncertainty}
& 75.74$\pm$0.98
& 82.34$\pm$0.02
& 82.92$\pm$0.09
& 2.54$\pm$0.74 \\

& & IUAV-IP~\cite{nahiduzzaman2026uncertainty}
& 59.41$\pm$0.25
& \underline{82.58$\pm$0.05}
& 82.47$\pm$0.05
& 3.89$\pm$0.05 \\





& & ReCAP (without pruning)
& \textbf{83.20$\pm$0.00}
& 82.04$\pm$1.11
& \textbf{83.47$\pm$0.95}
& 1.82$\pm$0.01 \\

& & ReCAP (with pruning)
& \textbf{83.20$\pm$0.00}
& 82.04$\pm$1.11
& \textbf{83.47$\pm$0.95}
& \underline{1.68$\pm$0.01} \\

\rowcolor{gray!15}
& & \textit{GISS-U}
& 96.53$\pm$0.00
& 96.44$\pm$0.29
& 96.45$\pm$0.29
& 1.16$\pm$0.01 \\

\midrule

\multirow{11}{*}{BrEaST}
& \multirow{3}{*}{Static}
& CBM~\cite{koh2020concept}
& 25.49$\pm$0.99
& 73.02$\pm$0.82
& 74.51$\pm$0.96
& 7.00$\pm$0.00 \\

& & ProbCBM~\cite{kim2023probabilistic}
& 25.49$\pm$1.10
& 75.32$\pm$0.92
& 76.47$\pm$0.54
& 7.00$\pm$0.00 \\

& & Evi-CEM~\cite{gao2024evidential}
& 29.51$\pm$0.15
& 68.26$\pm$0.24
& 72.84$\pm$0.32
& 7.00$\pm$0.00 \\

\cmidrule(lr){2-7}

& \multirow{8}{*}{Sequential}
& Joint-AFA~\cite{shim2018joint}
& \underline{72.23$\pm$0.83}
& 74.21$\pm$0.23
& 76.41$\pm$0.22
& 3.12$\pm$1.01 \\

& & Generative IP~\cite{chattopadhyay2022interpretable}
& 47.06$\pm$0.00
& 75.32$\pm$0.00
& 76.47$\pm$0.00
& 3.87$\pm$0.02 \\

& & V-IP~\cite{chattopadhyayvariational}
& 25.49$\pm$1.22
& 63.95$\pm$0.13
& 66.66$\pm$0.97
& 6.72$\pm$0.98 \\

& & EUAV-IP~\cite{nahiduzzaman2026uncertainty}
& 37.25$\pm$1.92
& \textbf{77.16$\pm$0.98}
& \underline{78.43$\pm$0.79}
& 5.27$\pm$0.92 \\

& & IUAV-IP~\cite{nahiduzzaman2026uncertainty}
& 43.14$\pm$1.35
& 74.84$\pm$1.21
& 76.47$\pm$1.04
& 4.65$\pm$1.24 \\





& & ReCAP (without pruning)
& 68.00$\pm$0.00
& \underline{76.40$\pm$2.22}
& \textbf{79.33$\pm$2.31}
& \underline{2.77$\pm$0.01} \\

& & ReCAP (with pruning)
& \textbf{73.33$\pm$1.16}
& \underline{76.40$\pm$2.22}
& \textbf{79.33$\pm$2.31}
& \textbf{1.76$\pm$0.09} \\

\rowcolor{gray!15}
& & \textit{GISS-U}
& 94.12$\pm$0.00
& 97.92$\pm$0.00
& 98.04$\pm$0.00
& 0.65$\pm$0.03 \\

\midrule

\multirow{11}{*}{PH2}
& \multirow{3}{*}{Static}
& CBM~\cite{koh2020concept}
& 35.00$\pm$0.55
& \underline{92.81$\pm$0.05}
& \underline{94.50$\pm$0.37}
& 8.00$\pm$0.00 \\

& & ProbCBM~\cite{kim2023probabilistic}
& 40.00$\pm$0.34
& 87.04$\pm$0.24
& 87.50$\pm$1.94
& 8.00$\pm$0.00 \\

& & Evi-CEM~\cite{gao2024evidential}
& 27.50$\pm$0.84
& 90.19$\pm$0.24
& \underline{94.50$\pm$0.23}
& 8.00$\pm$0.00 \\

\cmidrule(lr){2-7}

& \multirow{8}{*}{Sequential}
& Joint-AFA~\cite{shim2018joint}
& 32.50$\pm$1.02
& 90.45$\pm$0.32
& \underline{94.50$\pm$0.99}
& 7.50$\pm$0.49 \\

& & Generative IP~\cite{chattopadhyay2022interpretable}
& 55.83$\pm$2.89
& 89.71$\pm$3.08
& 92.50$\pm$2.50
& \underline{3.75$\pm$0.17} \\

& & V-IP~\cite{chattopadhyayvariational}
& 37.50$\pm$1.13
& 92.80$\pm$0.61
& \underline{94.50$\pm$0.74}
& 6.38$\pm$0.98 \\

& & EUAV-IP~\cite{nahiduzzaman2026uncertainty}
& 55.00$\pm$0.72
& 84.38$\pm$1.05
& 90.00$\pm$0.92
& 5.03$\pm$0.93 \\

& & IUAV-IP~\cite{nahiduzzaman2026uncertainty}
& 35.00$\pm$0.87
& \underline{92.81$\pm$0.05}
& \underline{94.50$\pm$0.35}
& 6.22$\pm$0.34 \\





& & ReCAP (without pruning)
& \underline{65.00$\pm$0.00}
& \textbf{92.83$\pm$0.00}
& \textbf{95.00$\pm$0.00}
& 3.82$\pm$0.15 \\

& & ReCAP (with pruning)
& \textbf{80.83$\pm$1.44}
& \textbf{92.83$\pm$0.00}
& \textbf{95.00$\pm$0.00}
& \textbf{1.68$\pm$0.07} \\

\rowcolor{gray!15}
& & \textit{GISS-U}
& 79.17$\pm$3.82
& 92.83$\pm$0.00
& 95.00$\pm$0.00
& 1.68$\pm$0.12 \\

\midrule

\multirow{11}{*}{OAI}
& \multirow{3}{*}{Static}
& CBM~\cite{koh2020concept}
& 65.79$\pm$0.23
& 59.80$\pm$0.34
& \textbf{74.90$\pm$0.74}
& 10.00$\pm$0.00 \\

& & ProbCBM~\cite{kim2023probabilistic}
& 46.84$\pm$0.72
& 51.85$\pm$0.45
& 72.27$\pm$0.36
& 10.00$\pm$0.00 \\

& & Evi-CEM~\cite{gao2024evidential}
& 35.17$\pm$0.89
& 55.31$\pm$0.37
& 73.38$\pm$0.54
& 10.00$\pm$0.00 \\

\cmidrule(lr){2-7}

& \multirow{8}{*}{Sequential}
& Joint-AFA~\cite{shim2018joint}
& 65.63$\pm$0.71
& 48.25$\pm$1.12
& 67.35$\pm$1.75
& 9.69$\pm$0.15 \\

& & Generative IP~\cite{chattopadhyay2022interpretable}
& 70.02$\pm$2.06
& 59.51$\pm$0.57
& 72.07$\pm$0.42
& 8.30$\pm$0.27 \\

& & V-IP~\cite{chattopadhyayvariational}
& 66.53$\pm$0.14
& 54.84$\pm$0.54
& 73.99$\pm$0.83
& 8.66$\pm$0.12 \\

& & EUAV-IP~\cite{nahiduzzaman2026uncertainty}
& 69.90$\pm$0.32
& \underline{59.81$\pm$0.47}
& 74.31$\pm$0.21
& 8.12$\pm$1.01 \\

& & IUAV-IP~\cite{nahiduzzaman2026uncertainty}
& 68.01$\pm$1.01
& 54.59$\pm$0.56
& 74.16$\pm$0.73
& 8.72$\pm$0.29 \\





& & ReCAP (without pruning)
& \underline{88.25$\pm$0.22}
& \textbf{60.21$\pm$0.29}
& \underline{74.83$\pm$0.36}
& \underline{5.36$\pm$0.20} \\

& & ReCAP (with pruning)
& \textbf{90.42$\pm$0.51}
& \textbf{60.21$\pm$0.29}
& \underline{74.83$\pm$0.36}
& \textbf{2.86$\pm$0.21} \\

\rowcolor{gray!15}
& & \textit{GISS-U}
& 98.39$\pm$0.13
& 82.92$\pm$0.07
& 93.57$\pm$0.05
& 1.71$\pm$0.21 \\

\midrule

\multirow{11}{*}{WBCatt}
& \multirow{3}{*}{Static}
& CBM~\cite{koh2020concept}
& 67.05$\pm$0.14
& 93.64$\pm$0.04
& 95.64$\pm$0.04
& 29.00$\pm$0.00 \\

& & ProbCBM~\cite{kim2023probabilistic}
& 47.67$\pm$0.87
& \underline{95.51$\pm$0.04}
& \underline{96.32$\pm$0.03}
& 29.00$\pm$0.00 \\

& & Evi-CEM~\cite{gao2024evidential}
& 70.25$\pm$0.23
& 94.32$\pm$0.04
& 94.54$\pm$0.03
& 29.00$\pm$0.00 \\

\cmidrule(lr){2-7}

& \multirow{8}{*}{Sequential}
& Joint-AFA~\cite{shim2018joint}
& 76.80$\pm$0.39
& 92.00$\pm$0.05
& 94.27$\pm$0.07
& 7.73$\pm$1.02 \\

& & Generative IP~\cite{chattopadhyay2022interpretable}
& 92.51$\pm$0.06
& 92.40$\pm$0.46
& 94.85$\pm$0.28
& 3.96$\pm$0.18 \\

& & V-IP~\cite{chattopadhyayvariational}
& 64.21$\pm$0.21
& 93.40$\pm$0.02
& 95.58$\pm$0.05
& 10.05$\pm$1.04 \\

& & EUAV-IP~\cite{nahiduzzaman2026uncertainty}
& 82.28$\pm$0.21
& 93.42$\pm$0.07
& 95.51$\pm$0.06
& 6.56$\pm$1.76 \\

& & IUAV-IP~\cite{nahiduzzaman2026uncertainty}
& 83.31$\pm$0.28
& 93.02$\pm$0.04
& 95.29$\pm$0.05
& 6.55$\pm$1.11 \\





& & ReCAP (without pruning)
& \underline{93.07$\pm$3.63}
& \textbf{95.52$\pm$0.16}
& \textbf{96.49$\pm$0.08}
& \underline{3.81$\pm$0.77} \\

& & ReCAP (with pruning)
& \textbf{95.77$\pm$0.85}
& \textbf{95.52$\pm$0.16}
& \textbf{96.49$\pm$0.08}
& \textbf{2.15$\pm$0.13} \\

\rowcolor{gray!15}
& & \textit{GISS-U}
& 93.27$\pm$1.46
& 96.71$\pm$0.15
& 97.29$\pm$0.20
& 2.27$\pm$0.62 \\

\bottomrule
\end{tabular}
\end{table*}

\subsection{Uncertainty-Based Selective Concept Reliability}
\label{sec:supp_selective_reliability}
UAV-IPs~\cite{nahiduzzaman2026uncertainty}
demonstrated that uncertainty estimates are informative
of concept-prediction correctness, with higher uncertainty
generally associated with incorrect concept predictions.
Here, we further investigate the empirical reliability
of concept predictions under uncertainty-based abstention
across five medical imaging datasets.

For each concept $m$, a prediction is considered admissible
when its estimated uncertainty satisfies
$U_m \leq \kappa_m$, where $\kappa_m$ is the
concept-specific uncertainty threshold.
Otherwise, the CP returns an abstention
($\bot$). We evaluate the raw concept-prediction error
rate before thresholding, the error rate among admissible
predictions, and the proportion of predictions retained
after thresholding.

As shown in Table~\ref{tab:selective_reliability},
uncertainty-based filtering consistently reduces
concept-prediction errors across all five datasets.
For instance, the error rate decreases from
23.25\% to 9.60\% on BrEaST, from 18.32\% to
10.99\% on DDR, and from 8.23\% to 2.74\% on
WBCatt. The proportion of admissible predictions
varies across datasets, reflecting differences in
concept-prediction uncertainty and the corresponding
thresholds. 

\begin{table}[t]
\centering
\small
\caption{
\textbf{Concept-prediction reliability under uncertainty-based abstention.}
Raw Error denotes the concept-prediction error rate before
uncertainty filtering, while Adm. Error denotes the error
rate among predictions satisfying $U_m \leq \kappa_m$.
Admissible indicates the percentage of concept predictions
retained after uncertainty filtering.
}
\label{tab:selective_reliability}

\resizebox{\columnwidth}{!}{%
\begin{tabular}{lccc}
\toprule
\textbf{Dataset} &
\textbf{Raw Error (\%)} &
\textbf{Adm. Error (\%)} &
\textbf{Admissible (\%)} \\
\midrule
DDR    & 18.32 & 10.99 & 70.36 \\
BrEaST & 23.25 &  9.60 & 59.58 \\
PH2    & 20.62 & 13.70 & 84.38 \\
OAI    &  6.80 &  2.38 & 79.82 \\
WBCatt &  8.23 &  2.74 & 75.86 \\
\bottomrule
\end{tabular}%
}
\end{table}

\subsection{Reliability under Increasing Concept Acquisition}
\label{sec:supp_fixed_budget}

To assess whether ReCAP's reliability arises from fewer
queries or more reliable concept acquisition, we compare
ReCAP with IUAV-IP, V-IP, and Generative IP on WBCatt,
which contains 29 concepts. We evaluate the first
$k=2,\ldots,10$ acquired concepts along each method's
natural acquisition trajectory, considering only samples
with at least $k$ acquisitions.

\begin{figure*}[t]
    \centering
    \includegraphics[width=\textwidth]
    {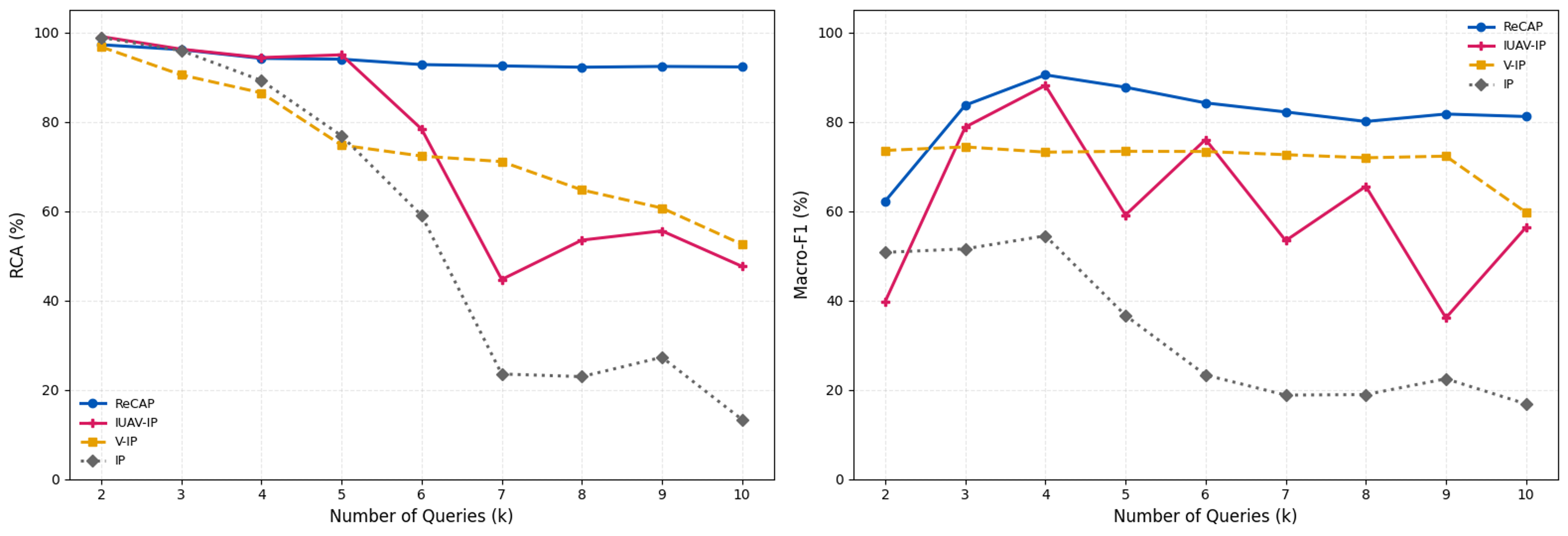}
    \caption{
    \textbf{Reliability and diagnostic performance under
    increasing concept acquisition on WBCatt.}
    Comparison of ReCAP, IUAV-IP, V-IP, and Generative IP
    for $k=2,\ldots,10$.
    Left: trajectory-level RCA.
    Right: diagnostic Macro-F1.
    Only trajectories containing at least $k$ acquisitions
    are evaluated. ReCAP maintains high reliability across
    longer acquisition trajectories, indicating that its
    advantage stems from more reliable concept acquisition
    rather than simply fewer queries.
    }
    \label{fig:supp_wbcatt_fixed_k}
\end{figure*}

As shown in Fig.~\ref{fig:supp_wbcatt_fixed_k},
ReCAP maintains RCA above 90\% across all acquisition
lengths, whereas competing methods exhibit substantial
reliability degradation as more concepts are acquired.
Importantly, this advantage persists when comparing
trajectories at the same query budget, demonstrating that
ReCAP's higher reliability is not merely a consequence
of acquiring fewer concepts, but reflects more reliable
concept selection. ReCAP also maintains competitive
diagnostic Macro-F1, exceeding 80\% from $k=3$ onward.

\subsection{Ground-Truth-Informed State Search}
\label{sec:supp_giss}

To examine the reliability of concept histories selected
with access to the ground-truth diagnosis, we consider
Ground-Truth-Informed State Search (GISS-U) using the
frozen ReCAP diagnosis classifier.
Unlike sequential acquisition, which selects concepts
based on the currently observed history, GISS-U
constructs candidate concept histories and selects
a terminal state using the true class label.

\paragraph{Candidate state construction.}
For each test image $x$, we first identify the
uncertainty-admissible concept set
$\mathcal{A}(x)=\{m:\sigma_m(x)\leq\kappa_m\}$,
where $\kappa_m$ is the uncertainty threshold
for concept $m$. Concepts exceeding their
thresholds are excluded from the search.
For each admissible concept, we use the predicted
answer $\hat c_m(x)$ from the concept predictor,
without replacing it with the ground-truth
concept annotation.

We construct candidate states by selecting subsets
$S\subseteq\mathcal{A}(x)$ containing at most
$B$ concepts, where $B$ is the acquisition budget.
Each subset defines a history
$H_S=\{(m,\hat c_m(x)):m\in S\}$.
The candidate space is therefore
\begin{equation}
\mathcal{S}(x)=
\{S\subseteq\mathcal{A}(x):|S|\leq B\}.
\end{equation}
Since the histories contain predicted concept
answers, incorrect concept predictions remain
possible even when the ground-truth diagnosis
is available during state selection.

\paragraph{State search and selection.}
We evaluate each candidate history using the
frozen diagnosis classifier $p_\eta(y\mid H_S)$.
For each state, we compute the predicted diagnosis
$\hat y(H_S)$, the Gibbs criterion
$G_\eta(H_S)$, the number of acquired concepts,
and the posterior probability assigned to the
ground-truth class.

Candidate states are ranked lexicographically
according to four criteria: (i) whether the
predicted diagnosis matches the ground-truth
class, (ii) whether the Gibbs stopping criterion
is satisfied, (iii) the number of acquired
concepts, preferring shorter histories, and
(iv) the posterior probability of the true
diagnostic class. Formally,
\begin{equation}
\begin{aligned}
S^*=\operatorname*{lex\,argmax}_{S\in\mathcal{S}(x)}
\big(&\mathbb{I}[\hat{y}(H_S)=y],\\
&\mathbb{I}[G_\eta(H_S)\leq\delta'],\\
&-|S|,\ p_\eta(y\mid H_S)\big),
\end{aligned}
\label{eq:giss_objective}
\end{equation}
where $y$ is the ground-truth diagnosis and
$\delta'$ is the Gibbs stopping tolerance.

The lexicographic ordering ensures that diagnostic
correctness takes precedence over all subsequent
criteria. Among states with the same diagnostic
correctness, those satisfying the stopping criterion
are preferred. If these conditions are identical,
the state containing fewer concepts is selected.
The true-class posterior is used only to
distinguish states tied on the preceding criteria.

\paragraph{Search implementation.}
We enumerate all candidate subsets when the
number of possible states does not exceed 4,096.
Otherwise, we sample 128 candidate states using
stratified sampling across acquisition depths,
allowing histories with different numbers of
concepts to be considered.
Each candidate is evaluated independently by
the frozen diagnosis classifier, and the
highest-ranked state according to
Eq.~\eqref{eq:giss_objective} is retained.
No additional classifier training or sequential
acquisition policy is involved in this search.

\paragraph{Interpretation.}
GISS-U uses privileged information only for
selecting the terminal concept history: the
ground-truth diagnosis determines which candidate
states are preferred. It does not correct
incorrect concept predictions or explicitly
optimise concept reliability. Consequently,
a selected state may contain incorrect concepts
if those concepts do not prevent a correct
diagnosis or satisfaction of the stopping
criterion.

GISS-U therefore provides a ground-truth-informed
diagnostic reference rather than a theoretical
upper bound on RCA or a deployable acquisition
policy. Moreover, when candidate states are
sampled rather than exhaustively enumerated,
the selected state is optimal only among the
evaluated candidates. Diagnosis-preserving
pruning may consequently achieve higher RCA
by removing incorrect or redundant concepts
that remain in the GISS-U selected history.

\subsection{Qualitative Analysis of Concept Acquisition}
\label{sec:supp_qualitative_analysis}

\begin{figure*}[t]
    \centering
    \includegraphics[width=\textwidth]{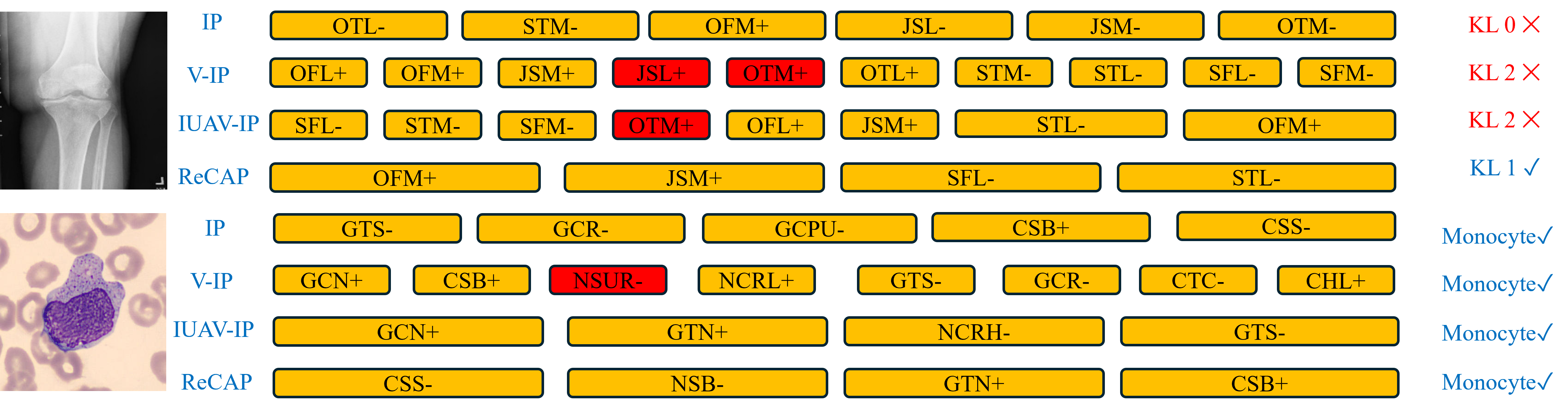}
    \caption{
    \textbf{Qualitative comparison of concept acquisition
    trajectories on OAI (top) and WBCatt (bottom).}
    The figure compares Generative IP, V-IP, IUAV-IP and ReCAP.
    Concepts are shown in acquisition order, with
    $+$ and $-$ indicating positive and negative acquired
    answers, respectively. Yellow boxes indicate correct
    concept acquisitions, whereas red boxes indicate
    incorrect acquisitions relative to ground-truth
    concept annotations.
    On OAI, IP acquires six concepts without errors but
    predicts KL grade 0. V-IP and IUAV-IP acquire ten
    and eight concepts, respectively, with two and one
    incorrect acquisitions, and both predict KL grade 2.
    ReCAP acquires four concepts without errors and
    correctly predicts KL grade 1.
    On WBCatt, all four methods correctly predict the
    monocyte class. IP, IUAV-IP, and ReCAP acquire five,
    four, and four concepts without errors, respectively,
    while V-IP acquires eight concepts with one incorrect
    acquisition.
    Full concept names are reported in
    Table~\ref{tab:dataset_details}.
    }
    \label{fig:qualitative_trajectories}
\end{figure*}

Figure~\ref{fig:qualitative_trajectories} presents qualitative
comparisons of concept acquisition trajectories produced by
Generative IP, V-IP, IUAV-IP, and ReCAP on representative samples
from OAI and WBCatt. Each trajectory shows the acquired concepts
in their selection order, together with their observed positive
($+$) or negative ($-$) answers. Incorrect concept acquisitions
are highlighted in red, while correct acquisitions are
highlighted in yellow.

\noindent\textbf{OAI.}
On OAI, Generative IP acquires six concepts without incorrect
acquisitions but predicts KL grade 0. V-IP acquires ten concepts,
including two incorrect acquisitions, and predicts KL grade 2.
Similarly, IUAV-IP acquires eight concepts with one incorrect
acquisition and predicts KL grade 2. In contrast, ReCAP
acquires only four concepts without incorrect acquisitions
and correctly predicts KL grade 1.
This example illustrates that avoiding incorrect concept
acquisitions does not necessarily guarantee a correct diagnosis,
as the final prediction also depends on how the acquired
concepts are used for classification. ReCAP achieves both
reliable concept acquisition and correct diagnosis with
fewer queries.

\noindent\textbf{WBCatt.}
On WBCatt, all four methods correctly classify the sample
as a monocyte. Generative IP acquires five concepts without
incorrect acquisitions, while V-IP acquires eight concepts,
including one incorrect acquisition (NSUR$-$).
IUAV-IP and ReCAP each acquire four concepts without
incorrect acquisitions. This example demonstrates that
correct diagnostic predictions do not necessarily imply
reliable concept acquisition. Although all methods reach
the correct diagnosis, their acquisition trajectories
differ in both reliability and query efficiency.

Overall, these examples complement the quantitative results
in Table~\ref{tab:main_full_results} by illustrating the
distinction between diagnostic correctness, trajectory-level
concept reliability, and query efficiency.

\subsection{Effect of Acquisition Strategy}
\label{sec:acquisition_ablation}

We investigate the effect of acquisition strategy and
objective on concept reliability and explanation compactness.
We compare amortised acquisition methods
(V-IP, EUAV-IP, and IUAV-IP) with search-based methods
(Generative IP and ReCAP).
Table~\ref{tab:strategy_comparison} reports the average
RCA and $Q_{\mathrm{avg}}$ across five medical imaging
datasets, with each dataset contributing equally.

\begin{table}[t]
\centering
\caption{
\textbf{Effect of acquisition strategy.}
Unweighted cross-dataset averages of Reliable Concept
Acquisition Rate (RCA) and $Q_{\mathrm{avg}}$ across
five medical imaging datasets.}
\label{tab:strategy_comparison}

\scriptsize
\setlength{\tabcolsep}{3pt}
\renewcommand{\arraystretch}{1.05}

\resizebox{\columnwidth}{!}{%
\begin{tabular}{@{}lllcc@{}}
\toprule
\textbf{Method}
& \textbf{Selection}
& \textbf{Objective}
& \textbf{RCA}$\uparrow$
& $\mathbf{Q}_{\mathrm{avg}}\downarrow$ \\
\midrule

V-IP
& Amort.
& Info. gain
& 50.03
& 7.16 \\

EUAV-IP
& Amort.
& Info. gain
& 64.03
& 5.50 \\

IUAV-IP
& Amort.
& Info. gain
& 57.77
& 6.01 \\

Generative IP
& Search
& Info. gain
& 65.36
& 4.71 \\

\textbf{ReCAP (Ours)}
& Search
& Eq.~13
& \textbf{84.71}
& \textbf{2.02} \\

\bottomrule
\end{tabular}%
}
\end{table}

Among amortised methods, EUAV-IP achieves the highest
RCA (64.03\%) with 5.50 queries, while the search-based
Generative IP achieves 65.36\% RCA with 4.71 queries.
ReCAP further improves RCA to 84.71\%, exceeding
Generative IP by 19.35 percentage points while retaining
only 2.02 concepts after terminal pruning.
These results demonstrate the effectiveness of ReCAP
in selecting reliable concept evidence and producing
compact explanations.

\subsection{Stopping Behaviour}
\label{sec:stopping_analysis}

We analyze the sensitivity of ReCAP to the stopping tolerance
$\delta'$. Increasing $\delta'$ encourages earlier stopping,
reducing the number of acquired concepts and potentially
improving RCA by limiting exposure to incorrect concept
predictions. However, premature stopping may reduce
diagnostic performance. We therefore select $\delta'$
by jointly considering RCA, $Q_{\mathrm{avg}}$,
macro-F1, and accuracy.

Table~\ref{tab:delta_sensitivity} shows that the stopping
tolerance has a limited effect on DDR, which has a
relatively small concept space. RCA remains unchanged
at $91.04\%$ across the evaluated tolerances. This suggests
that stopping behaviour is relatively stable for this
dataset. We select $\delta'=0.05$, as the evaluated
tolerances produce nearly identical acquisition reliability
and query costs, indicating limited sensitivity to this
parameter on DDR.

In contrast, OAI, which has a larger concept space,
shows greater sensitivity to the stopping tolerance.
Increasing $\delta'$ from $0.05$ to $0.20$ reduces
$Q_{\mathrm{avg}}$ from $5.65$ to $4.37$ and improves
RCA from $86.66\%$ to $88.62\%$. However, these
improvements come at the cost of diagnostic performance.
Increasing $\delta'$ from $0.10$ to $0.20$ reduces
macro-F1 from $51.50\%$ to $50.65\%$ and accuracy
from $72.16\%$ to $71.29\%$. We therefore select
$\delta'=0.10$ to balance acquisition reliability,
query efficiency, and diagnostic performance.

Overall, these results suggest that stopping tolerance
has less influence when the concept space is small,
whereas larger concept spaces may require more careful
selection of the stopping criterion. In particular,
higher RCA and fewer queries do not necessarily lead
to better diagnostic performance. We therefore prioritize
a balance between reliability and diagnostic accuracy
rather than selecting the tolerance that maximizes RCA
or minimizes the number of queries. The same
validation-based selection procedure is applied to
the remaining datasets.

\begin{table}[t]
\centering
\small
\caption{
\textbf{Validation-set sensitivity to the stopping tolerance
$\delta'$.}
Results are mean $\pm$ sample standard deviation over
three seeds. Bold indicates the selected operating point.
}
\label{tab:delta_sensitivity}
\setlength{\tabcolsep}{3.5pt}
\begin{tabular}{lccc}
\toprule
\textbf{Dataset}
& $\boldsymbol{\delta'}$
& \textbf{RCA}$\uparrow$
& $\mathbf{Q}_{\mathrm{avg}}\downarrow$ \\
\midrule

\multirow{5}{*}{DDR}
& 0.010
& $91.04\pm0.00$
& $1.444\pm0.000$ \\
& 0.020
& $91.04\pm0.00$
& $1.444\pm0.000$ \\
& 0.030
& $91.04\pm0.00$
& $1.444\pm0.000$ \\
& 0.040
& $91.04\pm0.00$
& $1.442\pm0.002$ \\
& \textbf{0.050}
& $\mathbf{91.04\pm0.00}$
& $\mathbf{1.442\pm0.002}$ \\
\midrule

\multirow{5}{*}{OAI}
& 0.050
& $86.66\pm0.69$
& $5.65\pm0.22$ \\
& 0.075
& $86.79\pm0.57$
& $5.30\pm0.03$ \\
& \textbf{0.100}
& $\mathbf{86.87\pm0.57}$
& $\mathbf{5.20\pm0.02}$ \\
& 0.150
& $87.94\pm0.00$
& $4.59\pm0.03$ \\
& 0.200
& $88.62\pm0.19$
& $4.37\pm0.03$ \\
\bottomrule
\end{tabular}
\end{table}

\subsection{Synthetic Experiments}
\label{appendix:synthetic}

\paragraph{Why simulate.}
On real data the true concepts, the image posterior $P(y\mid x)$ and the
optimal policy are unknown, so incorrect assertions can only be estimated,
Proposition~\ref{prop:decomp} cannot be checked and
Theorem~\ref{thm:budget} cannot be compared with the optimum. The simulation
makes all three available. Eight binary concepts keep the optimal policy
computable by dynamic programming over the $3^8$ histories.

\paragraph{Data generation.}
Each case has eight binary true concepts $c^\star$ and a binary label. Two
concepts $c_1,c_2$ are uniform and independent and set the clean label
$y_0=c_1\oplus c_2$. Six further concepts are drawn independently given $y_0$,
with $P(c_m=1\mid y_0=1)=a_m$ and $P(c_m=1\mid y_0=0)=b_m$: three \emph{rare}
concepts with $(a_m,b_m)=(0.35,0)$, present in 17.5\% of cases and decisive
for $y_0=1$ when present, and three \emph{weak} cues with $(0.25,0.05)$,
present in 15\% of cases. The observed label flips $y_0$ with probability
0.05. Alone, $c_1$ and $c_2$ carry no information about the label, each rare
concept 0.20 bits and each weak cue 0.061 bits; the pair carries one bit
jointly.
The concept predictor errs in an instance-specific, correlated way. Each case
has a latent difficulty $z_i\sim\mathcal{N}(0,1)$, and concept $m$ is misread,
$\hat c_{im}=c^\star_{im}\oplus E_{im}$, with
$P(E_{im}=1)=\operatorname{sigmoid}(\alpha+\lambda z_i)$, independently across
concepts given $z_i$. The offset $\alpha$ is shared by all concepts and set
for a mean error rate of 5\% (so no concept is more reliable than another),
and $\lambda=1.5$ (mean pairwise error correlation 0.13). The uncertainty is
a noisy reading of the same difficulty,
$\sigma_{im}=\operatorname{sigmoid}(\alpha+\lambda z_i+\nu_{im})$ with
$\nu_{im}\sim\mathcal{N}(0,1)$ (AUROC 0.79 for detecting errors). With
abstention, $\kappa$ is the 80\% quantile of $\sigma$ on the training cases,
and a query returns $\hat c_{im}$ if $\sigma_{im}\le\kappa$ and $\bot$
otherwise (Equation~\eqref{eq:admissible}); a refusal costs a query, asserts
nothing and enters the history. One query in five is refused, and the error
rate of answered concepts falls from 4.9\% to 2.6\%.

\paragraph{Probabilities.}
Each run uses 6{,}000 training and 3{,}000 test cases. \emph{Counted}:
$p(y\mid\mathcal{H})$ and $p(a\mid m,\mathcal{H})$ are frequencies over the
training cases consistent with $\mathcal{H}$, with Laplace smoothing
(pseudo-count 1). \emph{Learned}: heads as in Sec.~\ref{sec:method}, each a
multilayer perceptron with two hidden layers of 64 ReLU units, trained with
Adam (learning rate $10^{-3}$, weight decay $10^{-4}$, batch 128). The input
encodes each concept as $-1$, $+1$, $\bot$ or unqueried; the answer head
outputs an $M\times|\mathcal{A}|$ table with
$\mathcal{A}=\{-1,+1,\bot\}$ (or $\{-1,+1\}$ without abstention). Stage 1
trains the label head on complete histories (10 epochs); stage 2 trains both
heads on random partial histories, drawing for every case and epoch a subset
of concepts of size uniform in $\{0,\dots,M\}$, with the answer loss over
unacquired concepts only (40 epochs); stage 3 fine-tunes both heads for 150
steps, each on 64 trajectories generated with fixed parameters by the rule
being evaluated.

\paragraph{Policies and evaluation.}
ReCAP selects by Equation~\eqref{eq:greedy_rule}; information gain selects the
concept of highest mutual information with the label given the history, as in
IP~\cite{chattopadhyay2022interpretable}; random selects uniformly. All rules
score every answer in $\mathcal{A}$, consider every unqueried concept, and,
within a variant, use the same probabilities. The budget $k$ counts queries,
refusals included; Gibbs stopping ends at $G(\mathcal{H})\le\delta$,
$\delta\in\{0,0.05,\dots,0.4\}$, or when every concept is queried. The
diagnosis is the MAP label. Pruning applies Equation~\eqref{eq:inert_approx}
with $\xi=0.05$ as in Algorithm~\ref{alg:greedy}, removing the first inert
concept in index order until none remains; refusals are never displayed, and
RCA is computed over the displayed concepts. Results average three seeds (120,
130, 140; 90, 100, 110 for the sweeps). Table~\ref{tab:synthetic8} lists the
fixed-budget accuracies.

\begin{table*}[t]
\centering
\footnotesize
\setlength{\tabcolsep}{4.5pt}
\caption{Diagnostic accuracy against query budget, with abstention ($\kappa$;
solid and dashed curves of Fig.~\ref{fig:synthetic}a) and without. At $k=8$
every concept has been queried.}
\label{tab:synthetic8}
\begin{tabular}{lllcccccccc}
\toprule
& Heads & Rule & $k=1$ & $k=2$ & $k=3$ & $k=4$ & $k=5$ & $k=6$ & $k=7$ & $k=8$ \\
\midrule
\multirow{6}{*}{\rotatebox{90}{With $\kappa$}}
 & \multirow{3}{*}{Learned} & Random & 0.570 & 0.627 & 0.683 & 0.730 & 0.764 & 0.789 & 0.814 & 0.836 \\
 & & Info.\ gain & 0.621 & 0.698 & 0.750 & 0.760 & 0.763 & 0.787 & 0.807 & 0.838 \\
 & & ReCAP & 0.511 & 0.799 & 0.813 & 0.820 & 0.826 & 0.830 & 0.836 & 0.838 \\
\cmidrule{2-11}
 & \multirow{3}{*}{Counted} & Random & 0.567 & 0.631 & 0.685 & 0.729 & 0.756 & 0.760 & 0.766 & 0.752 \\
 & & Info.\ gain & 0.619 & 0.701 & 0.749 & 0.753 & 0.748 & 0.734 & 0.724 & 0.752 \\
 & & ReCAP & 0.497 & 0.792 & 0.806 & 0.814 & 0.817 & 0.808 & 0.783 & 0.752 \\
\midrule
\multirow{6}{*}{\rotatebox{90}{Without $\kappa$}}
 & \multirow{3}{*}{Learned} & Random & 0.577 & 0.651 & 0.718 & 0.760 & 0.795 & 0.827 & 0.856 & 0.882 \\
 & & Info.\ gain & 0.641 & 0.728 & 0.778 & 0.781 & 0.783 & 0.791 & 0.820 & 0.882 \\
 & & ReCAP & 0.504 & 0.875 & 0.875 & 0.877 & 0.880 & 0.881 & 0.881 & 0.883 \\
\cmidrule{2-11}
 & \multirow{3}{*}{Counted} & Random & 0.578 & 0.667 & 0.708 & 0.767 & 0.791 & 0.825 & 0.855 & 0.872 \\
 & & Info.\ gain & 0.652 & 0.735 & 0.779 & 0.779 & 0.785 & 0.787 & 0.804 & 0.872 \\
 & & ReCAP & 0.494 & 0.866 & 0.866 & 0.866 & 0.866 & 0.869 & 0.872 & 0.872 \\
\bottomrule
\end{tabular}
\end{table*}

\paragraph{Counted against learned probabilities.}
Without abstention the two agree to within 0.016. With abstention they agree
to within 0.014 up to $k=4$, after which the counted probabilities degrade to
0.752 with every concept queried, against 0.838 for the learned heads. Each
queried concept then has three outcomes, so a complete history is one of
$3^8=6{,}561$ patterns; in one seed only 1{,}547 occur in training, 15\% of
complete test histories match no training case (their smoothed posterior is
uniform), and the counted diagnosis is correct for 30\% of these against 89\%
for histories matched ten or more times. The learned heads share parameters
across histories and do not degrade. Since all rules reach the same history at
$k=8$, the dip is a property of the counted estimator, not of the rules.

\begin{figure*}[t]
\centering
\includegraphics[width=0.87\textwidth]{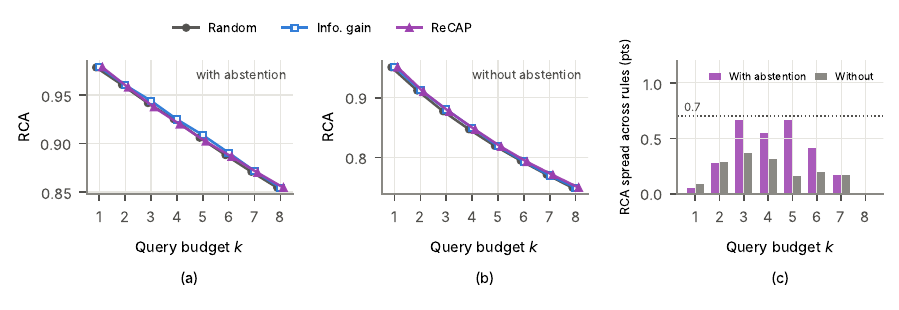}
\caption{\textbf{RCA at a fixed budget} (learned heads). (a) RCA against the
budget $k$ with abstention ($k$ counts queries, refusals included). (b) The
same without abstention. (c) Spread of RCA across the three rules at each
budget (maximum minus minimum, in points); dotted: 0.7 points. Markers are
offset horizontally for visibility.}
\label{fig:synthetic_fixed_rca}
\end{figure*}

\paragraph{Reliability.}
Every concept has the same error rate, so at a fixed budget all rules assert
the same number of concepts (to within 0.02) and attain the same RCA (to
within 0.8 points counted, 0.7 learned; Fig.~\ref{fig:synthetic_fixed_rca}).
Under Gibbs stopping without abstention (counted, $\delta=0.30$), ReCAP's
error rate per asserted concept is 4.8\% against 4.2\% for information gain,
so its higher RCA (0.910 vs.\ 0.823) comes from asserting fewer concepts
(2.00 vs.\ 5.15).

\paragraph{Guarantee checks.}
Without abstention and under the counted probabilities, the optimal $k$-query
policy is computed by dynamic programming for each rule's own objective:
$f=R(\varnothing)-\mathbb{E}[R(\mathcal{H})]$ for ReCAP and the reduction of
terminal entropy for information gain (Fig.~\ref{fig:synthetic}e). For the
learned variants the rule selects with learned heads and its value is
evaluated under the counted probabilities. In a control without the pair, and
in the interaction sweep below at $\gamma=0$, greedy information gain reaches
at least 98\% of its optimum; its worst fraction over $k=2$--$7$ falls to 0.71
at $\gamma=2$ and 0.43 at $\gamma=4$, while greedy ReCAP stays at the optimum.
For Proposition~\ref{prop:decomp}, ReCAP runs with Gibbs stopping (counted, no
abstention), and the image posterior $P(y\mid x)=P(y\mid c^\star)$ is counted
on $2\times10^5$ further cases. Across $\delta$ the MAP error (0.128--0.134)
stays below $\mathbb{E}[G(\mathcal{H})]$ (0.200--0.213), and the Brier
discrepancy matches $\mathbb{E}[G(\mathcal{H})]-\mathbb{E}[G_x]$ to within
0.01, the residual reflecting finite training data. Label noise and predictor
errors keep $G(\mathcal{H})$ above about 0.20 for most cases, so small
tolerances are rarely reached and the bound then holds at
$\mathbb{E}[G(\mathcal{H})]$ rather than at $\delta$.

\begin{figure*}[t]
\centering
\includegraphics[width=\textwidth]{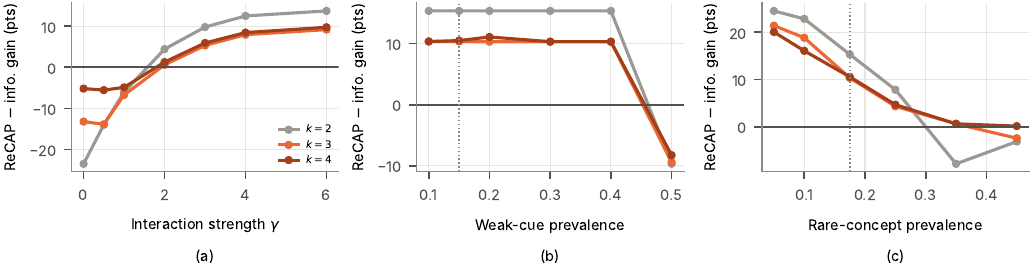}
\caption{\textbf{Operating regimes}, one property varied at a time, no
abstention. ReCAP minus information-gain accuracy against (a) the interaction
strength $\gamma$ ($\gamma=0$: the pair is uninformative), (b) the prevalence
of the weak cues at fixed information (0.064 bits; 0.5 is balanced), and
(c) the prevalence of the rare concepts, for $k=2$--$4$. Dotted lines mark the
scenario of Fig.~\ref{fig:synthetic}.}
\label{fig:synthetic_regimes}
\end{figure*}

\paragraph{Operating regimes.}
Each sweep varies one property, without abstention and with counted
probabilities (Fig.~\ref{fig:synthetic_regimes}).
\emph{Interaction strength.} The pair is replaced by $s_1,s_2$ uniform in
$\{-1,+1\}$ with
$\operatorname{logit}P(y_0=1\mid s)=\gamma\,s_1s_2+\mu\,(s_1+s_2)$, balanced
by subsampling, $\gamma\in\{0,0.5,1,2,3,4,6\}$ with $\mu=0$, and
$\mu\in\{0,0.25,\dots,1\}$ with $\gamma=3$. ReCAP's first query is a pair
member at every $\gamma$, including $\gamma=0$ where the pair is noise, so it
trails by 23.5 points at $k=2$ for $\gamma=0$, is level or slightly ahead at $\gamma=2$ and
leads by 5--14 points for $\gamma\ge3$ ($k=2$--$4$); $\mu=1$ reduces the lead
at $\gamma=3$ from 9.8 to 6.6 points at $k=2$.
\emph{Weak-cue prevalence} $p\in\{0.10,0.15,0.20,0.30,0.40,0.50\}$, with
$a_m-b_m$ solved so that each cue carries 0.064 bits. ReCAP leads by
10--15 points for $p\le0.40$ and trails by 8--10 points at the balanced
$p=0.5$, where it acquires the cues before the pair.
\emph{Rare-concept prevalence} $a/2$ with $(a_m,b_m)=(a,0)$,
$a\in\{0.1,\dots,0.9\}$. ReCAP's lead at $k=2$ falls from 24.5 points at
prevalence 0.05 to 7.8 at 0.25; from 0.35 information gain is ahead at $k=2$
and the rules are within 2.4 points for $k\ge3$.
\emph{Predictor quality} (Fig.~\ref{fig:synthetic}b): mean error in
$\{0,0.05,0.10,0.20\}$, accuracy at $k=3$.

\subsection{Empirical Analysis of the Guarantees: Details}
\label{appendix:empirical_guarantees}

This section gives the construction behind Sec.~\ref{sec:empirical_guarantees}.

\paragraph{Empirical prior and objective.}
The evaluation split of OAI contains $N=1203$ images with labels $y_i$ and CP
confidences $\boldsymbol\mu_i\in[0,1]^M$, $M=10$. Admissibility is disabled, so
answers are binary:
$\hat c_{i,m}=2\,\mathds{1}[\mu_{i,m}\geq\tfrac12]-1$.
For a history $\mathcal H$, let $V(\mathcal H)$ index the images whose
thresholded vector agrees with every answer in $\mathcal H$, with
$N_{\mathcal H}=|V(\mathcal H)|$ and $N_{\mathcal H,y}$ the number of these
with label $y$. Then
\begin{equation}
\begin{aligned}
\hat p(\mathcal H)
&=\frac{N_{\mathcal H}}{N},
\qquad
\hat p(y\mid\mathcal H)
=\frac{N_{\mathcal H,y}}{N_{\mathcal H}},\\
\hat p(a\mid m,\mathcal H)
&=\frac{|\{i\in V(\mathcal H):\hat c_{i,m}=a\}|}
{N_{\mathcal H}}.
\end{aligned}
\end{equation}
and Eq.~\eqref{eq:residual} becomes
\begin{equation}
\begin{aligned}
\hat R(\mathcal H)
&=\frac{N_{\mathcal H}^2-\sum_y N_{\mathcal H,y}^2}{N^2},\\
\hat f(\mathcal H)
&=\hat R(\varnothing)-\hat R(\mathcal H).
\end{aligned}
\label{eq:supp_emp_residual}
\end{equation}
with $\hat R(\mathcal H)=0$ when $N_{\mathcal H}=0$. On OAI,
$\hat R(\varnothing)=0.6322$, which bounds $\hat f$ from above. Its numerical
closeness to $1-1/e$ is a coincidence. No learned head is used to compute
$\hat f$.

\paragraph{Exact adaptive OPT.}
Let $V^\star(\mathcal H,b)$ be the largest expected terminal objective
attainable from $\mathcal H$ with $b$ queries remaining:
\begin{equation}
\begin{aligned}
V^\star(\mathcal H,0)
&=\hat f(\mathcal H),\\
V^\star(\mathcal H,b)
&=\max_{m\notin\mathrm{dom}(\mathcal H)}
\sum_{a\in\{-1,+1\}}
\hat p(a\mid m,\mathcal H)\\
&\quad\cdot
V^\star\big(
\mathcal H\cup\{(m,a)\},b-1
\big).
\end{aligned}
\label{eq:supp_opt_dp}
\end{equation}
and $f_{\mathrm{OPT}}(k)=V^\star(\varnothing,k)$. The policy is adaptive: at
$k=2$, for example, the best second concept is chosen separately for each
answer to the first, and the first concept maximises the resulting expected
value. Histories are keyed by their queried indices and answer signs and
memoised, so each (state, remaining budget) pair is evaluated once. Because
the recursion optimises against the evaluation distribution itself, OPT is an
oracle reference, not a deployable policy.

\paragraph{Exact greedy.}
Greedy selects
$\arg\max_{m}\Delta(m\mid\mathcal H)$ with Eq.~\eqref{eq:gain_def} evaluated
under $\hat p$, equivalently Eq.~\eqref{eq:greedy_rule} with the empirical
answer probabilities and Gibbs errors (Eq.~\eqref{eq:gain_expanded}). It
recomputes gains after every observed answer, so it is adaptive rather than a
fixed ordering. Greedy and OPT coincide at $k=1$; for $k>1$ OPT can do better
by accounting for later queries. We report
$f_{\mathrm{Greedy}}(k)=\mathbb E_{\hat p}[\hat f(\mathcal H_k^{\mathrm{Greedy}})]$.

\paragraph{ReCAP.}
We use the trained heads $p_\psi$ and $p_\eta$ after trajectory fine-tuning,
frozen. At each step ReCAP selects
$\arg\min_m s_\eta(m\mid\mathcal H)$ (Sec.~\ref{subsec:algorithm}), observes
$a_t=\hat c_{i,m_t}$ and appends it, until $k$ concepts are acquired. Selection
uses no evaluation-set frequencies. The acquired histories are then scored with
the same empirical objective:
\begin{equation}
f_{\mathrm{ReCAP}}(k)=\frac1N\sum_{i=1}^N
\hat f\big(\mathcal H^{\mathrm{ReCAP}}_{i,k}\big),
\end{equation}
so $f_{\mathrm{ReCAP}}/f_{\mathrm{OPT}}$ and
$f_{\mathrm{Greedy}}/f_{\mathrm{OPT}}$ are ratios under a common objective.

\paragraph{Objective estimated by the learned heads (ReCAP-Learn).}
To test whether the heads also estimate the \emph{value} of $f$, we rescore the
same ReCAP trajectories
$\tau_{i,k}=((m_1,a_1),\dots,(m_k,a_k))$ with learned quantities only. The
history probability is estimated by the chain of answer-head predictions,
\begin{equation}
\tilde p_\psi(\tau_{i,k})=\prod_{t=1}^k
p_\psi(a_t\mid m_t,\mathcal H_{t-1}),
\end{equation}
accumulated in log space, and
\begin{equation}
\begin{aligned}
\tilde f(\tau_{i,k})
&= G_\eta(\varnothing)
-\tilde p_\psi(\tau_{i,k})^2
G_\eta(\mathcal H_{i,k}),\\
f_{\mathrm{ReCAP\text{-}Learn}}(k)
&= \frac{1}{N}\sum_i \tilde f(\tau_{i,k}).
\end{aligned}
\end{equation}
The ratio $f_{\mathrm{ReCAP\text{-}Learn}}/f_{\mathrm{OPT}}$ compares two
different objectives and is not an approximation ratio. The answer-head
conditionals need not be consistent with a single joint distribution, so
$\tilde p_\psi$ can depend on acquisition order. We use it only as a
diagnostic of the heads' calibration.

\begin{table*}[t]
\centering
\scriptsize
\setlength{\tabcolsep}{4pt}
\caption{\textbf{OAI fixed-budget comparison, all budgets.} OPT, Greedy and
ReCAP are scored by the empirical objective $\hat f$; ReCAP-Learn rescores
ReCAP's histories with the learned heads. G/OPT and R/OPT are ratios under a
common objective; R-Learn/OPT is a cross-objective diagnostic. Early stopping and pruning are disabled.}
\label{tab:greedy_optimal_full}
\begin{tabular}{rrrrrrrr}
\toprule
$k$ & OPT & Greedy & ReCAP & ReCAP-Learn & G/OPT & R/OPT & R-Learn/OPT\\
\midrule
1  & 0.503418 & 0.503418 & 0.503418 & 0.501224 & 1.000000 & 1.000000 & 0.995640\\
2  & 0.591133 & 0.590514 & 0.590055 & 0.575033 & 0.998953 & 0.998177 & 0.972764\\
3  & 0.617526 & 0.616025 & 0.610723 & 0.592303 & 0.997570 & 0.988983 & 0.959156\\
4  & 0.623170 & 0.622841 & 0.619133 & 0.597200 & 0.999473 & 0.993523 & 0.958326\\
5  & 0.625672 & 0.625529 & 0.625254 & 0.598352 & 0.999772 & 0.999331 & 0.956334\\
6  & 0.626415 & 0.626381 & 0.626389 & 0.598736 & 0.999945 & 0.999958 & 0.955813\\
7  & 0.626432 & 0.626421 & 0.626413 & 0.598776 & 0.999982 & 0.999971 & 0.955852\\
8  & 0.626434 & 0.626433 & 0.626423 & 0.598771 & 1.000000 & 0.999984 & 0.955841\\
9  & 0.626434 & 0.626434 & 0.626431 & 0.598763 & 1.000000 & 0.999996 & 0.955829\\
10 & 0.626434 & 0.626434 & 0.626434 & 0.598757 & 1.000000 & 1.000000 & 0.955818\\
\bottomrule
\end{tabular}
\end{table*}

\paragraph{Learned versus empirical objective.}
The learned estimate falls below the empirical value at every budget, from
$0.5012$ vs.\ $0.5034$ at $k=1$ to $0.5988$ vs.\ $0.6264$ at $k=10$. The gap
widens up to $k\approx5$ and then stays roughly constant. Possible sources are
a mismatch between the learned and empirical root Gibbs error, error in the
chained history probability. Even so, ReCAP's selections score within $1.1\%$ of OPT under the
empirical objective. Selection depends only on the ranking of
$s_\eta$ across candidates, which can be accurate when absolute values are
not.

\section{Limitations and Future Work}
\label{sec:limitations}

Although ReCAP improves the reliability and efficiency of sequential
concept acquisition, some limitations remain. First, ReCAP does
not introduce a new uncertainty quantification mechanism but instead
relies on existing uncertainty estimates to guide reliable concept
acquisition. Future work will investigate alternative uncertainty
quantification and calibration techniques to further improve acquisition
reliability. Second, our experiments primarily focus on medical imaging,
where reliable and interpretable predictions are particularly important.
Future work will extend the evaluation to natural image datasets and
larger-scale concept spaces to investigate the generalizability and
scalability of the proposed framework. Finally, evaluating trajectory-level
Reliable Concept Acquisition (RCA) requires ground-truth concept
annotations to determine whether acquired concepts are correct.
Obtaining such annotations can be expensive and time-consuming,
particularly for large-scale datasets. Future work will explore
alternative concept verification strategies, including vision-language
models (VLMs), human expert feedback, and other annotation-efficient
approaches, to reduce dependence on ground-truth concept annotations
and enable reliability assessment in more realistic settings.

\end{document}